\documentclass[nopreprintline,12pt]{elsarticle}

\usepackage[utf8]{inputenc} % Ensures UTF-8 encoding (optional in modern LaTeX)
\usepackage{amsmath} % For equations
\usepackage{amssymb} % For mathematical symbols
\usepackage{lineno} % For line numbers
\usepackage{subcaption} % For subfigures
\usepackage{tabularx} % For tabularx tables
\usepackage{multirow} % For multirow cells in tables
\usepackage{makecell} % For makecell in tables
\usepackage{rotating} % For sidewaystable
\usepackage{units} % For \unit command
\usepackage{gensymb} % For \textdegree
\usepackage{textcomp} % For \textmu
\usepackage{todonotes} % For \todo notes
\setuptodonotes{inline}
\begin{document}

\begin{frontmatter}



\title{Precision Robotic Spot-Spraying: Reducing Herbicide Use and Enhancing Environmental Outcomes in Sugarcane}


\author[inst1,inst4,corref{cor1}]{Mostafa Rahimi Azghadi}
\affiliation[inst1]{organization={College of Science and Engineering, James Cook University},%Department and Organization
            %addressline={1 James Cook Drive}, 
            %city={Townsville},
            %postcode={4811}, 
            state={QLD},
            country={Australia}}

\author[inst2]{Alex Olsen}
\author[inst2]{Jake Wood}
\author[inst1,inst4]{Alzayat Saleh}
\author[inst1]{Brendan Calvert}
\author[inst3]{Terry Granshaw}
\author[inst3]{Emilie Fillols}
\author[inst1,inst4,corref{cor2}]{Bronson Philippa}

\affiliation[inst4]{organization={Agriculture Technology and Adoption Centre, James Cook University},             
state={QLD},
country={Australia}}

\affiliation[inst2]{organization={AutoWeed Pty Ltd},%Department and Organization
%addressline={Address Two}, 
%postcode={22222}, 
state={QLD},
country={Australia}}

\affiliation[inst3]{organization={Sugar Research Australia},%Department and Organization
            %addressline={Address Two}, 
            %city={Townsville},
            %postcode={22222}, 
            state={QLD},
            country={Australia}}

\cortext[cor1]{mostafa.rahimiazghadi@jcu.edu.au}
\cortext[cor2]{bronson.philippa@jcu.edu.au}

\begin{abstract}
Precise robotic weed control plays an essential role in precision agriculture. It can help significantly reduce the environmental impact of herbicides while reducing weed management costs for farmers. In this paper, we demonstrate that a custom-designed robotic spot spraying tool based on computer vision and deep learning can significantly reduce herbicide usage on sugarcane farms. We present results from field trials that compare robotic spot spraying against industry-standard broadcast spraying, by measuring the weed control efficacy, the reduction in herbicide usage, and the water quality improvements in irrigation runoff. The average results across 25 hectares of field trials show that spot spraying on sugarcane farms is 97\% as effective as broadcast spraying and reduces herbicide usage by 35\%, proportionally to the weed density. For specific trial strips with lower weed pressure, spot spraying reduced herbicide usage by up to 65\%. Water quality measurements of irrigation-induced runoff, three to six days after spraying, showed reductions in the mean concentration and mean load of herbicides of 39\% and 54\%, respectively, compared to broadcast spraying. These promising results reveal the capability of spot spraying technology to reduce herbicide usage on sugarcane farms without impacting weed control and potentially providing sustained water quality benefits.
\end{abstract}




\end{frontmatter}


\section{Introduction}
\label{sec:intro}
Herbicides are a serious threat to non-target plants and animals because they are readily carried by water run-off from farmland into rivers, creeks, coastal and inshore areas. Consequently, herbicide usage has attracted significant public attention and pressure from regulators. A promising approach to reduce herbicide runoff is to use precision agriculture and digital technologies \cite{Allmendinger2022}. Examples of such technologies include the use of Unoccupied Aerial Vehicles (UAV) for precise weed mapping, non-chemical robotic weed management, and the use of UAVs \cite{Rai2022} and ground vehicles \cite{Calvert2021} for spot-spraying of weeds. Spot spraying allows for a much more efficient use of chemicals since herbicides can be sprayed only where required, and not onto the bare ground or non-target plants.

In many scenarios, herbicides must be applied during the growth of the crop. This requires the spot sprayer to discriminate between weed and crop plants, a situation which is called green-on-green detection. This differs from so-called green-on-brown detection, which targets all plants indiscriminately. For instance, a common green-on-brown detection technology is Near-Infrared (NIR) optical sensing, which has been implemented in systems such as WeedSeeker \cite{Trimble} and Weed-IT \cite{WeedIT}. These systems react to the NIR light emitted by plants during photosynthesis. As a result, these systems do not discriminate between plant types and therefore cannot separate crop from weed. Hence, they can only be applied in a green-on-brown scenario, such as in fallow paddocks. 

Other sensor technologies for weed detection include multi-spectral, hyper-spectral, 3D, and depth cameras, as well as stereo and ultrasonic sensors. The most commonplace sensor that can be used to detect weeds in a green-on-green scenario is a standard Red-Green-Blue (RGB) camera, which works in the visible light spectrum. Precise detection of weeds among surrounding green plants, which can have similar features, is a complex problem in Computer Vision (CV). Fortunately, with the advent of Deep Learning (DL) technology and Convolutional Neural Networks (CNN) in the past decade, weed detection with image processing has progressed significantly \cite{chen2022performance}. 

% In this paper, we describe the development of a ground-based spot-spraying tool for sugarcane, which we call AutoWeed. Our system is illustrated in Fig. \ref{fig:autoweed_system}. We developed the computer vision algorithms for weed detection and designed the spraying system to be retrofitted to existing farm machinery. We carried out field trails in the Burdekin region of Queensland, Australia, which is one of Australia's largest sugarcane growing regions. We conducted a comprehensive set of experiments to measure the real-world impact of spot spraying in terms of the reduction of herbicide used, the resulting weed knockdown efficacy, and the measured concentration of herbicide active ingredients in water runoff induced by irrigation.

\begin{figure}[!t]
    \centering
    \begin{subfigure}[b]{0.49\textwidth}
        \centering
        \includegraphics[width=\textwidth] {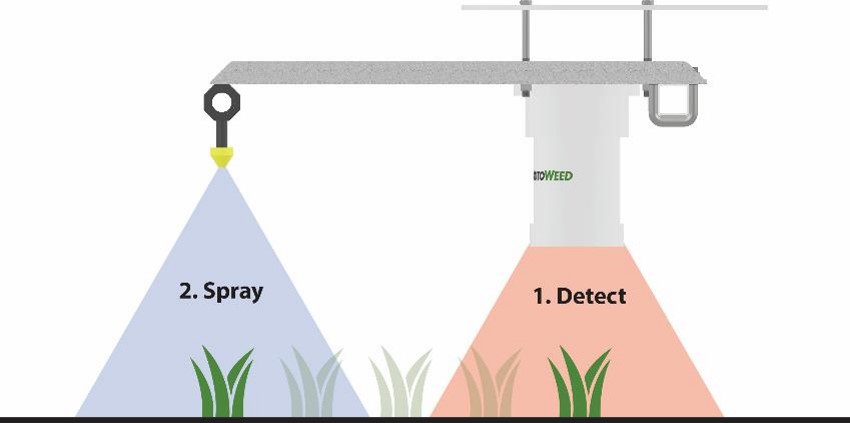}
        \caption{}
    \end{subfigure}
    \hfill
    \begin{subfigure}[b]{0.49\textwidth}
        \centering
        \includegraphics[width=\textwidth] {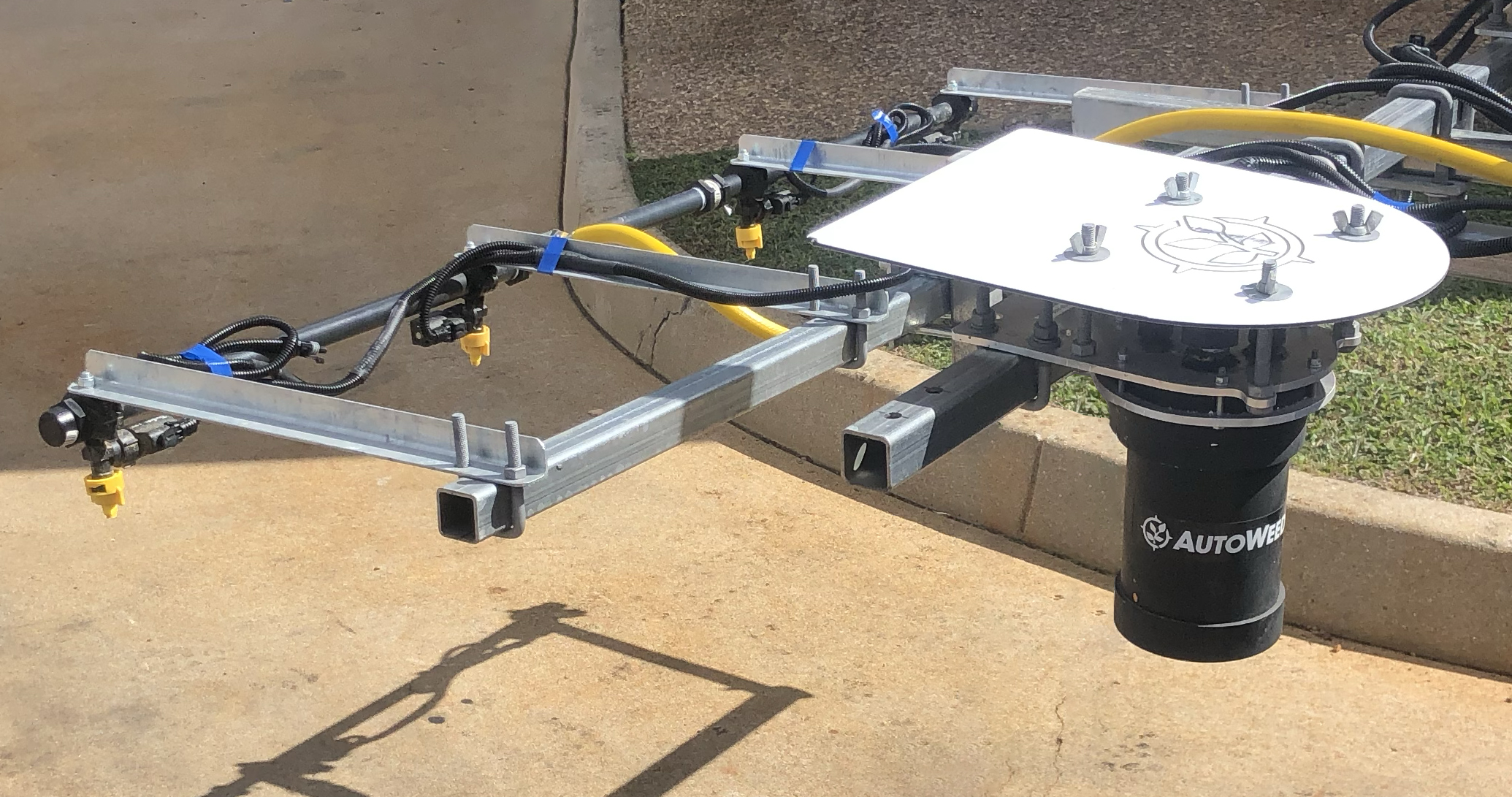}
        \caption{}
    \end{subfigure}
    \caption{(a) An illustration of the AutoWeed system where the AutoWeed unit detects the weed (1) and activates the sprayer (2). (b) A photo of the AutoWeed system mounted to a broadcast spraying boom.}
    \label{fig:autoweed_system}
\end{figure}

In this paper, we describe the development of a ground-based spot-spraying tool for sugarcane, which we call AutoWeed. Our system is illustrated in Fig. \ref{fig:autoweed_system}. We developed the computer vision algorithms for weed detection and designed the spraying system to be retrofitted to existing farm machinery. We carried out field trials in the Burdekin region of Queensland, Australia, which is one of Australia's largest sugarcane growing regions.

We conducted a comprehensive set of experiments to measure the real-world impact of spot spraying in terms of the reduction of herbicide used, the resulting weed knockdown efficacy, and the measured concentration of herbicide active ingredients in water runoff induced by irrigation. We hypothesised that: 

\begin{enumerate}
    \item  Spot spraying is more efficient in terms of herbicide usage compared to blanket spraying.
    \item  Spot spraying provides comparable weed knockdown efficacy compared to blanket spraying.
    \item  Spot spraying results in lower concentrations of herbicide active ingredients in water runoff compared to blanket spraying.
    \item  Spot spraying positively impacts crop yield compared to traditional blanket spraying methods.
\end{enumerate}

By addressing these hypotheses, we aim to provide a comprehensive evaluation of the efficacy and efficiency of the AutoWeed system.

The rest of this paper is organised as follows. Section \ref{Sec:Background} presents the background and related work. Section \ref{Sec:Methods} explains the methods used in this study including the DL detection algorithms, our collected datasets, our DL training and validation approaches, as well as the utilised water quality measurement protocols. In Section \ref{Sec:Results}, we provide on-ground trial results of our project including the weed killing efficacy, herbicide usage reduction, cost saving, and water quality improvements afforded by the technology, compared to current industry practice of broadcast (a.k.a. blanket) spraying. In Section \ref{Sec:Dis}, we provide discussions around the use of the technology and how its benefits can be expanded to other crops and farms. We also shed light on the future features and capabilities that can improve the uptake of robotic spot spraying to deliver a step change in water quality and environmental benefits. We conclude the paper in Section \ref{Sec:Conc}.  

\section{Background and Related Work}\label{Sec:Background}

\subsection{Deep learning and its application to precision agriculture}
Deep learning has grown to be a prevalent technology that has brought astonishing performance to computer vision. Deep learning-enabled CV algorithms are currently being applied in various precision agriculture domains.
% ranging from aquaculture \cite{Saleh2022} to marine \cite{Laradji2021} to environmental \cite{Jahanbakht2022}, and to medical sciences \cite{Azghadi2020,Mustafa2021}. 

Similarly, in the past few years, computer vision for precision agriculture and weed detection has significantly grown. For instance, Lu \emph{et al.} have conducted a survey of public datasets for computer vision tasks in precision agriculture using machine and deep learning \cite{Lu2020}. They have surveyed 10 studies for fruit detection, disease, damage and flower detection and counting, as well as yield prediction, using deep learning. 

Among the precision agriculture applications that Lu \emph{et al.} have surveyed, weed control has emerged as a major priority. They reported that there have been 15 publicly available ground and aerial image datasets collected in field using various sensors, such as RGB cameras, multi-spectral and multi-modal sensors, in the past 8 years. Several of these datasets such as \cite{Lameski2017} have been used in deep learning-based weed control tasks \cite{chen2022performance}. Here, weed control can be any use of computer vision and deep learning to segment, or simply classify  weeds for mapping or spraying.

Other studies have demonstrated the potential of deep learning for weed detection and control in precision agriculture. For instance, Jin \emph{et al.} \cite{Jin2022} developed a deep learning-based method for detecting herbicide weed control spectrum in turfgrass, which can be used in a machine vision-based autonomous spot-spraying system of smart sprayers. Similarly, Harders \emph{et al.} \cite{harders2022deep} proposed a deep learning approach for UAV-based weed detection in horticulture using edge processing and presented experimental results. In addition to detecting weeds, researchers have also explored the use of deep learning to generate synthetic weed images. Chen \emph{et al.} \cite{chen2023deep} applied diffusion probabilistic models to generate high-quality synthetic weed images based on transfer learning, while Divyanth \emph{et al.} \cite{divyanth2022image} aimed to reduce the effort needed to prepare large image datasets by creating artificial images of maize and common weeds through conditional Generative Adversarial Networks. Other researchers have focused on developing autonomous systems for weed detection and control. Patel \emph{et al.} \cite{patel2022design} described the design of an autonomous agricultural robot for real-time weed detection using CNN, while Gao \emph{et al.} \cite{gao2022transferring} developed a deep convolutional network that enables the prediction of both field and aerial images from UAVs for weed segmentation and mapping with only field images provided in the training phase.   Narayana \emph{et al.} \cite{narayana2023efficient} also developed an efficient real-time weed detection technique using deep learning. To facilitate the development of deep learning-based weed detection methods, Wang \emph{et al.} \cite{wang2022weed25} presented Weed25, a deep learning dataset for weed identification containing 14,035 images of 25 different weed species. Finally, Murad \emph{et al.} \cite{murad2023weed} conducted a systematic literature review on current state-of-the-art DL techniques for weed detection. 

In this paper, we use weed classification for spot-spraying. Classification, as opposed to segmentation, is an easier computer vision task that can be carried out very accurately using deep learning \cite{chen2022performance,Calvert2021}. The goal here is to identify if an image includes one or more weed(s) of interest.

\subsection{DL for weed spot-spraying}

Although many works have investigated and reported the potential of using CNNs for weed classification in different scenarios within on-ground \cite{Knoll2019} or UAV images, research works on in-field weed detection in conjunction with spot-spraying has not seen wide exploration. This is mainly due to the significant efforts required to perform efficient in-field trials outside a lab environment. However, recent advancements in deep learning have shown promising results in the detection and classification of weeds, which is crucial for the development of targeted spraying systems.

In one of the earliest studies \cite{Partel2019}, Partel \emph{et al.} developed a prototype spot-spraying system and simulated it using two scenarios. In the first scenario, a vegetable field was simulated using artificial weeds (targets) and artificial plants (non-targets), while in the second scenario, they applied their prototype to real plants. They also investigated the weed detection performance and system operations using two different Graphics Processing Units (GPUs) reporting the achieved accuracies. Finally, they utilized a GPS device integrated into their detection system to produce weed maps. Although this showcased the use of deep learning technology to perform spot-spraying of weeds, it was in a controlled and simulated environment.

In ref. \cite{Liu2021}, a variable rate chemical spraying system was partially trialled in the field at a low application speed of 3 km/hr using a prototype system controlling chemical usage in a strawberry crop. However, this was only a prototype design that did not report herbicide reduction achieved or the potential water quality benefits that could be obtained.

In ref. \cite{Du2022}, a CNN-based multi-class under-canopy weed control robotic unit named SAMBot was developed and trialled in a limited setting in flax fields with medium weed density. SAMBot was able to achieve an average weed classification accuracy of 90\%. It also showed reduced herbicide usage compared to a commercial sprayer. This study trialled the robot in only 15 m of fields and further results should be shown before the efficacy of the system could be fully demonstrated.

%In a recent study, Allmendinger \emph{et al.} reviewed different commercial technologies and prototypes for precision patch spraying and spot spraying \cite{Allmendinger2022}. They demonstrated that weed spot-spraying using DL has seen a lot of attention from the industry and several systems are currently being trialled and made commercially available. Examples of these products that use RGB cameras and DL include Bilberry, Blue River’s See and Spray, EcoPatch, and Greeneye. For a detailed review of these systems, please see \cite{Allmendinger2022}.

While significant advancements have been made in the application of deep learning for weed detection, the integration of these technologies into in-field weed spot-spraying systems remains under-explored. Future research should focus on conducting more extensive field trials and reporting on the practical benefits such as herbicide reduction and water quality improvements, which are the objectives of our paper.

\subsection{Current Status of Research on Target Spraying Robots}
Target spraying robots have become a pivotal technology in precision agriculture, significantly improving the accuracy and efficiency of herbicide application. 
In a recent study, Allmendinger \emph{et al.} reviewed different commercial technologies and prototypes for precision patch spraying and spot spraying \cite{Allmendinger2022}. They demonstrated that weed spot-spraying using DL has seen a lot of attention from the industry and several systems are currently being trialled and made commercially available. For a detailed review of these systems, please see \cite{Allmendinger2022}.
Key developments in target spraying robots include:\\

\noindent
\textbf{Autonomous Navigation and Detection Systems:} Modern target spraying robots are equipped with advanced navigation systems and real-time weed detection capabilities. For instance, autonomous robots like the one developed by \cite{abanay2022lidar} use GPS and LiDAR to navigate fields and identify weed locations accurately.

\noindent
\textbf{Integration of Deep Learning Models:} The use of deep learning models, such as CNNs, for weed detection has been a game-changer. Systems like those reviewed by \cite{rai2023applications} utilize DL models to distinguish between crop and weed species with high precision, enabling targeted herbicide application.

\noindent
\textbf{Field Trials and Practical Implementations:} Extensive field trials have been conducted to test the efficacy of target spraying robots under real-world conditions. Studies by \cite{sapkota2023towards} demonstrated significant herbicide savings and improved weed control efficacy in large-scale agricultural settings.

\noindent
\textbf{Commercialization and Industry Adoption:} Several commercial products have emerged from this research, including Bilberry \cite{Bilberry}, Blue River’s See and Spray \cite{BlueRiver}, and EcoPatch \cite{EcoPatch}. These systems are being adopted by farmers globally, driven by their potential to reduce chemical usage and enhance crop yields.

\section{Methods} \label{Sec:Methods}

\subsection{The AutoWeed system}
\begin{figure}[htbp]
    \centering
    \includegraphics[width=0.8\textwidth]{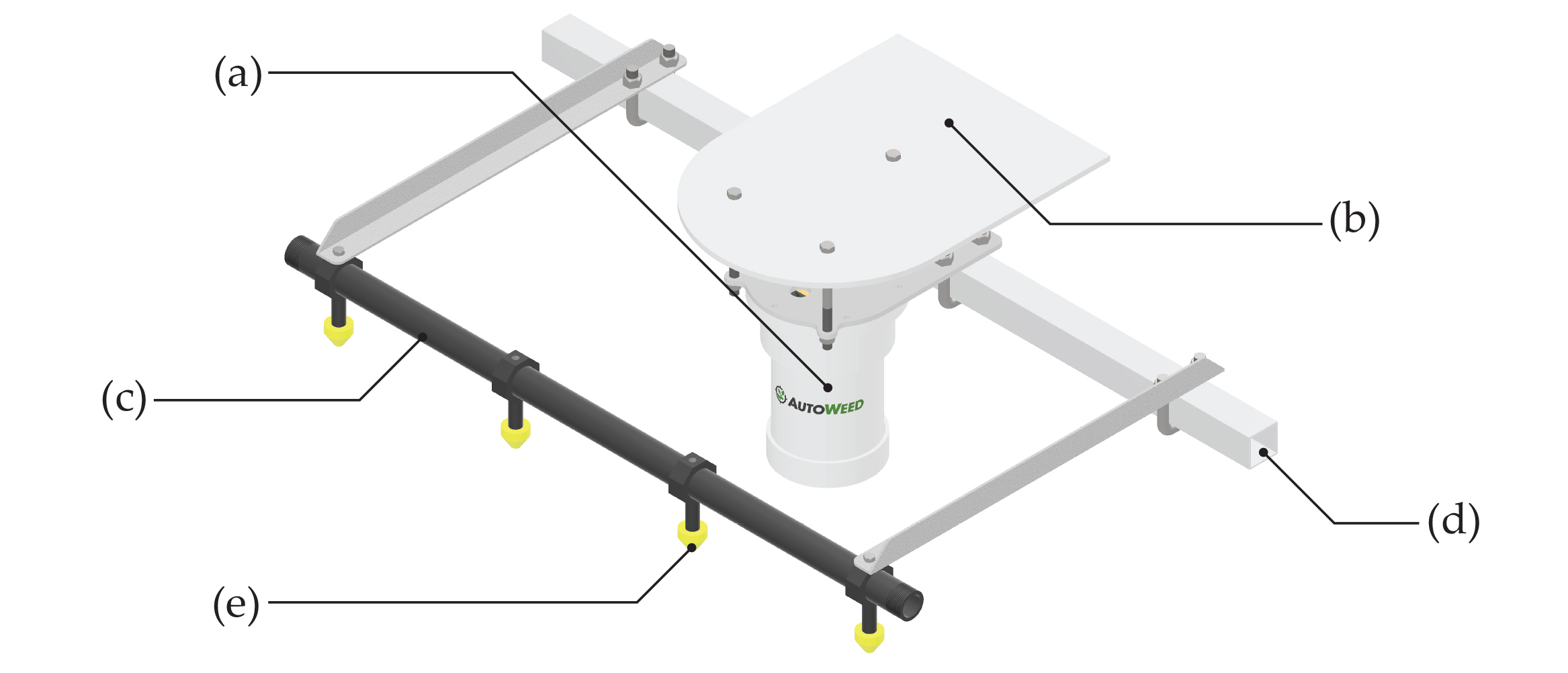}
    \caption{The AutoWeed weed detection and spraying boom mount design for a 1-metre boom section, including  (a) the AutoWeed detection unit including a machine vision camera, an NVIDIA Jetson GPU, and a custom solenoid sprayer board that can individually control up to four solenoids per camera; (b) a protective sun shade, (c) a 1`` wet boom, (d) a 40 $\times$ 40 mm steel hollow section frame, and (e) TeeJet solenoids and nozzle body adaptors.   }\label{fig:autoweed_detection_unit}
\end{figure}

The prototype AutoWeed units are shown in Figure \ref{fig:autoweed_detection_unit}.

These units were purpose-built for retrofitting to existing agricultural vehicles. Each unit comprises a machine vision camera, an NVIDIA Jetson embedded GPU processor, and a custom solenoid sprayer board that can individually control up to four solenoids per camera. The design is enclosed in a robust PVC housing attached to a mounting plate to protect the internals from the harsh environment. The units can be mounted to any 40 mm or 50 mm steel hollow section frame and are compatible with the commercially-available TeeJet boom and spray nozzle components.

\begin{figure}[htbp]
    \centering
    \begin{subfigure}[b]{0.48\textwidth}
        \centering
        \includegraphics[width=\textwidth]{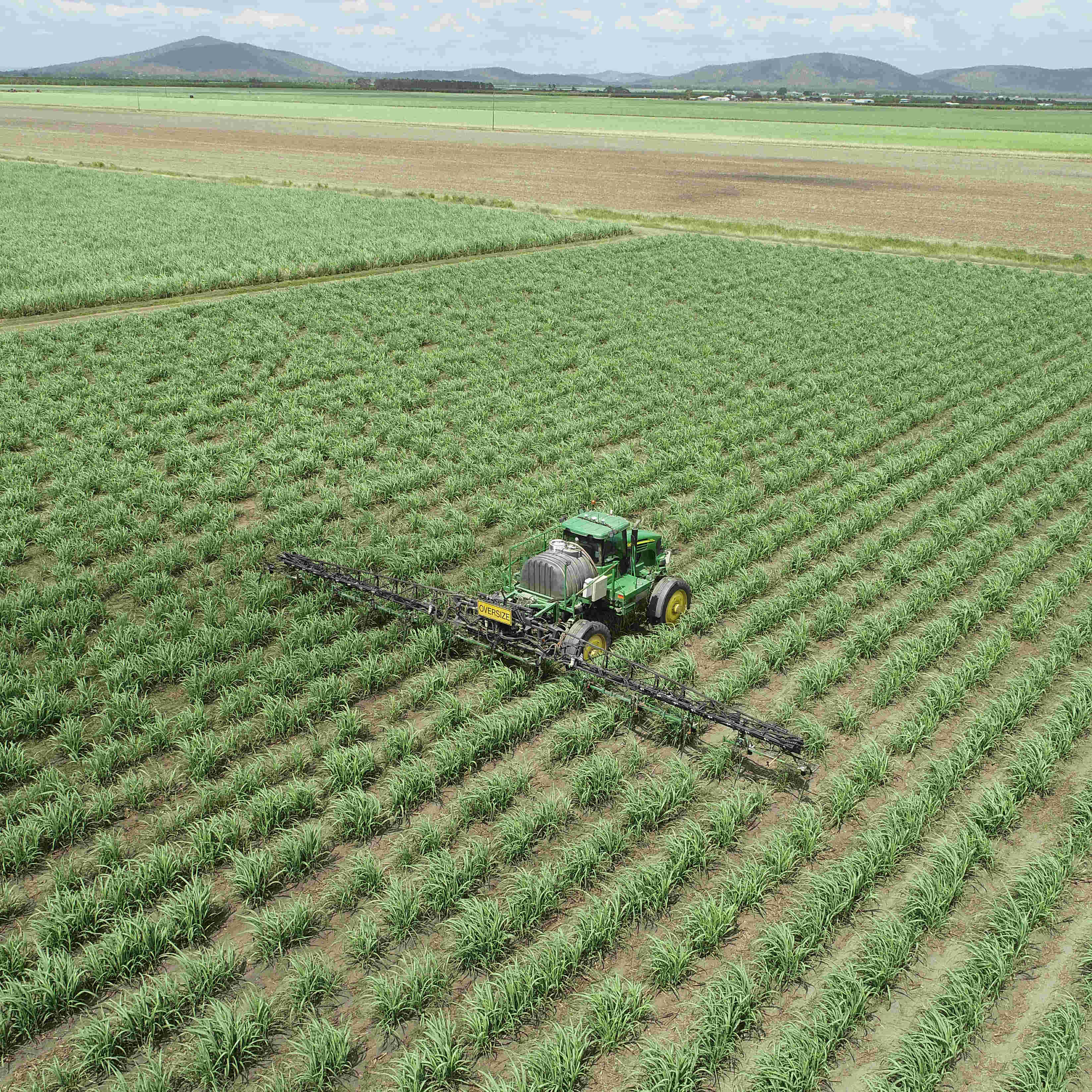}
        \caption{John Deere R4720 sprayer with AutoWeed}
        \label{subfig:3a}
    \end{subfigure}
    \hfill
    \begin{subfigure}[b]{0.48\textwidth}
        \centering
        \includegraphics[width=\textwidth]{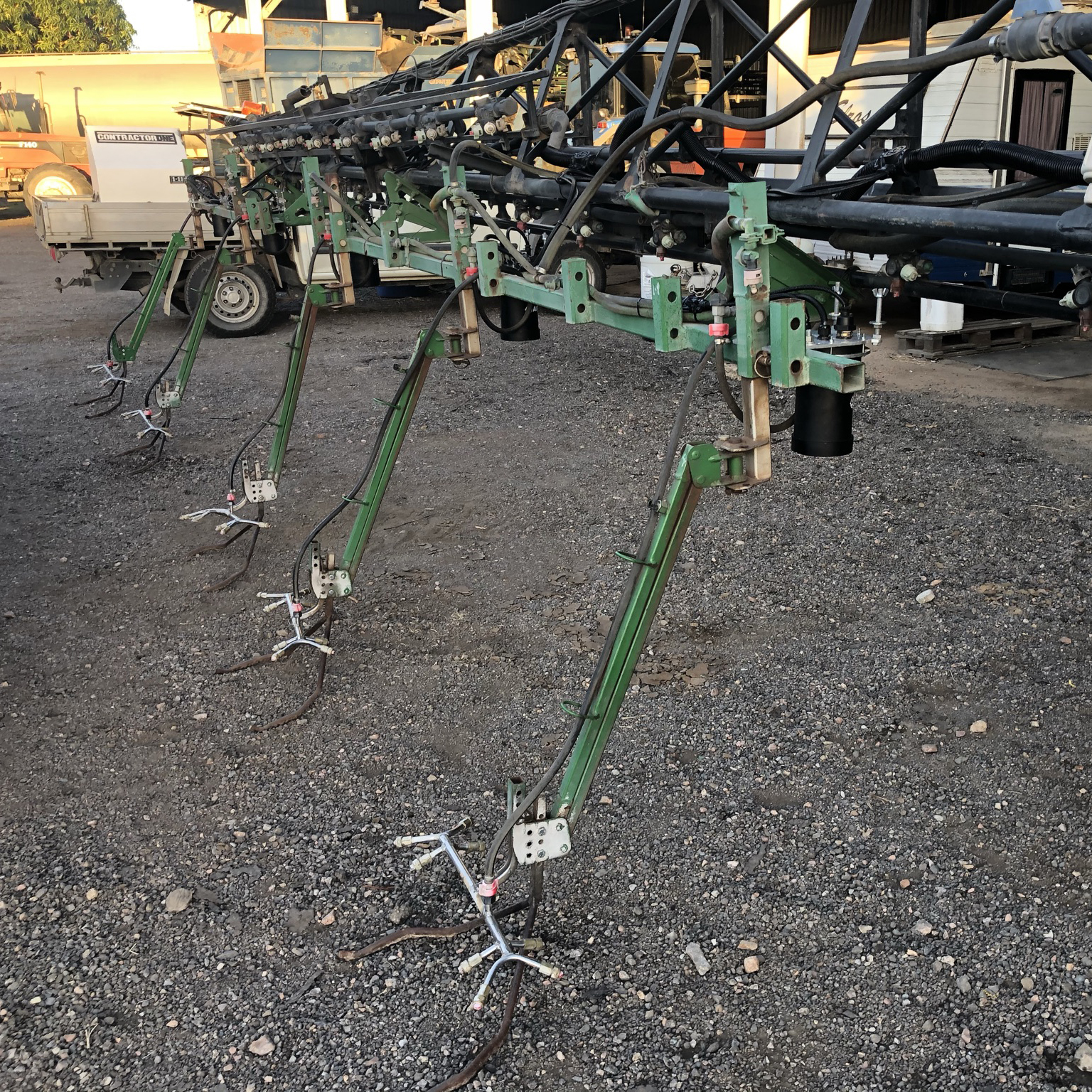}
        \caption{Detection units on Irvin legs}
        \label{subfig:3b}
    \end{subfigure}
    \par\bigskip % Replaces \vspace for consistent spacing
    \begin{subfigure}[b]{0.48\textwidth}
        \centering
        \includegraphics[width=\textwidth]{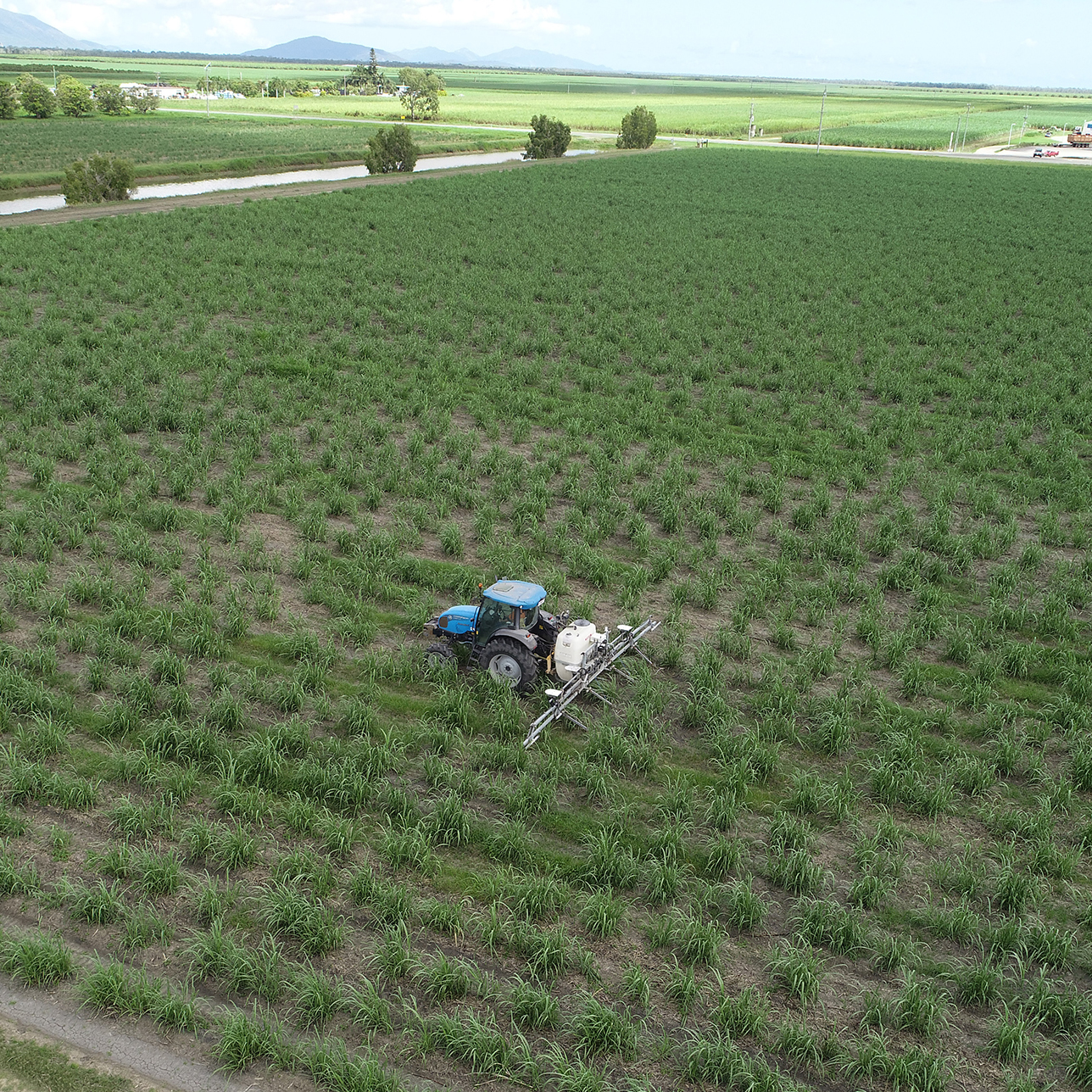}
        \caption{4-row sprayer for broadcast spraying}
        \label{subfig:4a}
    \end{subfigure}
    \hfill
    \begin{subfigure}[b]{0.48\textwidth}
        \centering
        \includegraphics[width=\textwidth]{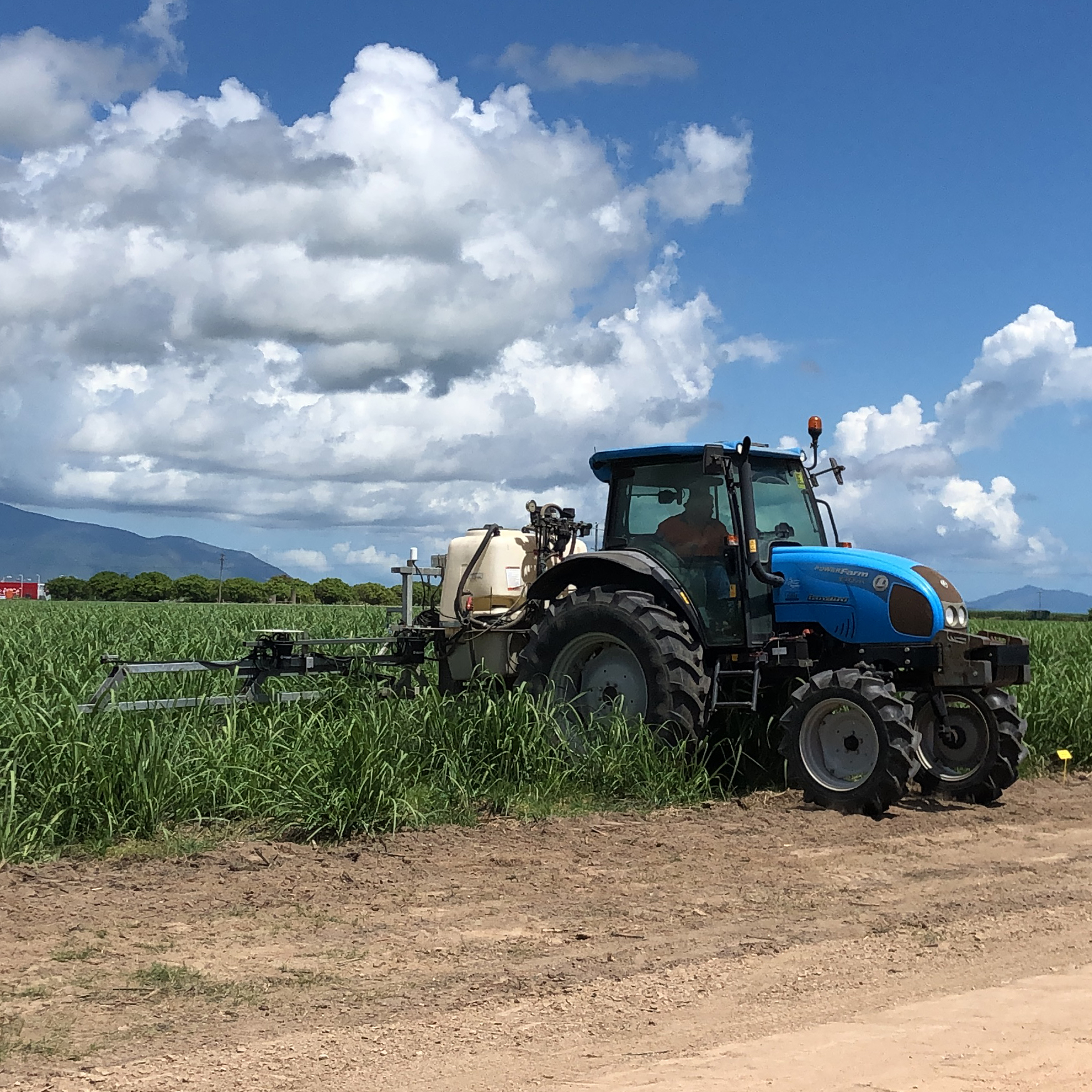}
        \caption{Detection units on spray boom frame}
        \label{subfig:4b}
    \end{subfigure}
    \caption{\color{black} 
    Illustration of the AutoWeed system retrofitted on different sprayers: (a) a 13-row high-rise John Deere R4720 self-propelled sprayer for Irvin leg spraying, with (b) detection units mounted in front of the Irvin legs between cane row centres; (c) a 4-row sprayer fitted for broadcast spraying, with (d) detection units mounted to the spray boom frame.
    \color{black} 
    }
    \label{fig:merged_sprayer}
\end{figure}

In this project, the AutoWeed detection and spraying prototypes were retrofitted for broadcast spraying and Irvin leg spraying on two spraying vehicles. The first was a large 13-row high-rise John Deere R4720 self-propelled sprayer (see Figure \ref{fig:merged_sprayer}, a and b). The second was a smaller 4-row three-point linkage boom sprayer towed behind a Landini tractor (as shown in Figure \ref{fig:merged_sprayer}, c and d). For both systems, the AutoWeed detection units were mounted in front of the sprayers and positioned between the sugarcane row centres.

\subsection{Weed image acquisition, analysis, and application technology}
The AutoWeed units utilise a machine vision camera pointing directly down to capture images of the field (see Fig. \ref{fig:autoweed_detection_unit}). 
The height of the camera in the detection unit, as shown in Fig. \ref{fig:autoweed_system}(b), is set based on the required field of view of the target crop row width. For sugarcane paddocks in the Burdekin region, the row width usually varies between 1.5 to 1.6 metres. To achieve a field of view that captured the full sugarcane row width, the camera lens was positioned at least 1m above the ground to ensure a horizontal field of view of at least 1.6 metres. We also utilised separately adjustable mounting for the camera and the spray nozzles so that the height of the spray nozzle to the target weeds could be adjusted without affecting the height of the camera.

Deep learning image classification models are trained using large labelled image datasets collected from the fields and weeds of interest. These models are then used in-field to analyse the images collected in real-time during spot-spraying and detect those containing weeds of interest to spot-spray them using our custom solenoid sprayer board that can
individually control up to four solenoids per detection unit (see Fig. \ref{fig:autoweed_detection_unit}e). %, following our previously reported methodology \cite{chen2022performance}. 
This process is explained in detail below.  

\subsubsection{The weed dataset}\label{DLdataset}
%Background on dataset use and couple of previous datasets such as DeepWeeds\\
One of the crucial first steps when developing a deep learning algorithm is collecting suitable datasets. The quantity, quality, and diversity of the data used to train the model directly impact its performance. For weed detection and spot-spraying, the collected dataset should contain diverse images of both crops and weeds in different growth stages, lighting conditions, and environments. The larger and more diverse the dataset, the better the model will be able to generalise and accurately identify weeds in new, unseen situations. It is also important to carefully label the data to ensure that the model is properly trained on the correct classifications \cite{chen2022performance}. In spot-spraying applications using deep learning, the image dataset is typically collected from a specific site and focuses on the weed(s) of interest, for which a dedicated model is trained. This method, known as site-specific weed control, is effective because it considers the unique conditions of each site, including the growth stage and the variety in shape and features of the same weeds, which can differ significantly between locations.

\begin{figure}[hbt!]
    \centering
    \includegraphics[width=0.8\textwidth] {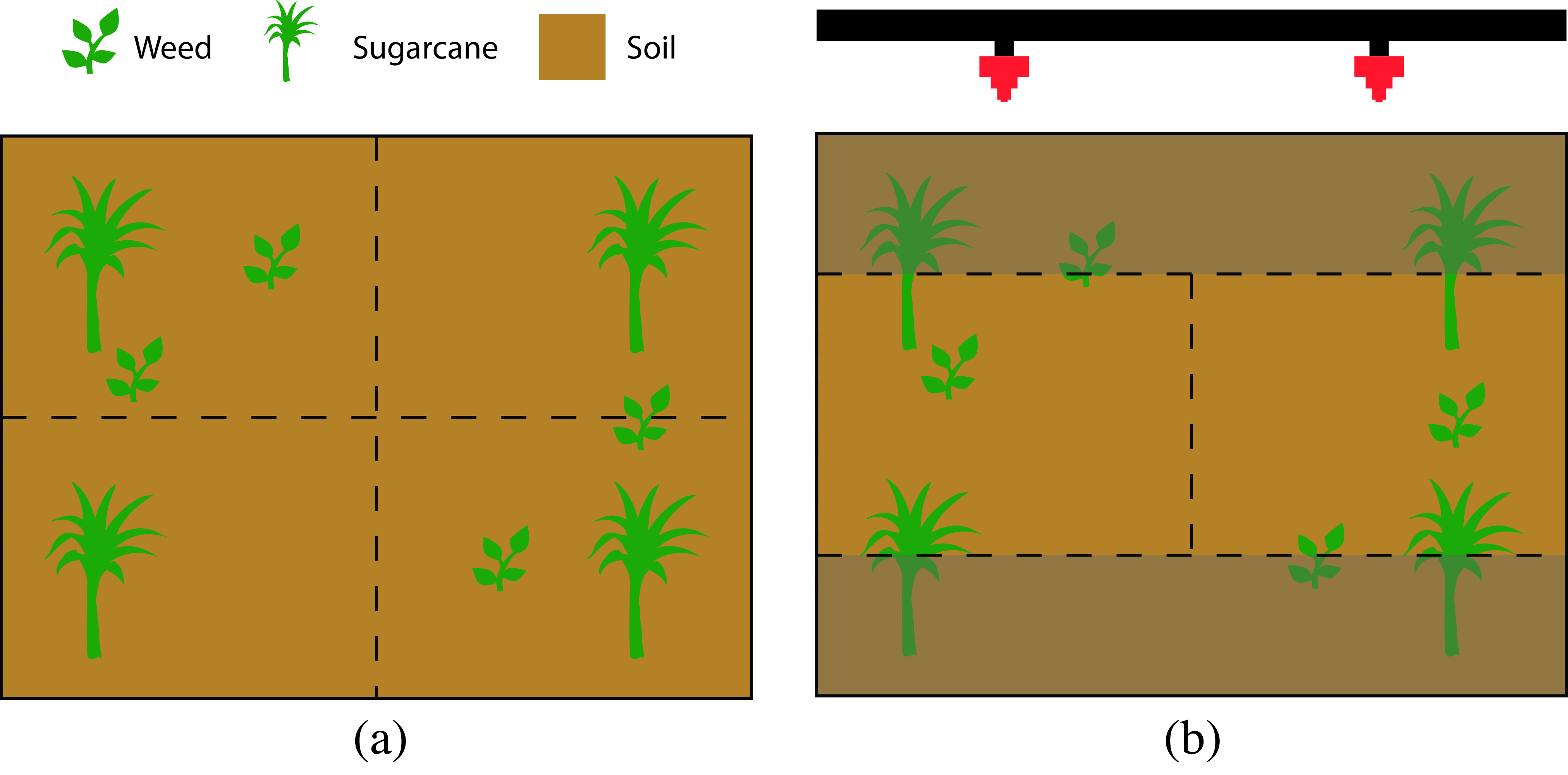}
    \caption{The classification approach for sugarcane where the camera is centred on the interrow and (a) images are split into four tiles for annotation and (b) the field of view is cropped to have two tiles of equal size in the centre of the frame for inference. For each tile, if a weed is detected the corresponding spray nozzle is activated.\label{fig:classification_approach}}
\end{figure}

%Discuss the dataset we collected and provide some sample pictures\\
For the project presented in this paper, we collected a total of 1,447,456 site-specific images before each of the respective six spray trials as described in Table \ref{tab:trial_data}. This data allowed training site-specific deep learning algorithms to detect target weeds for the different sugarcane paddocks. Data collection was performed in the Burdekin region of Queensland, Australia using an All-Terrain Vehicle (ATV) retrofitted with AutoWeed detection units that include a machine vision camera pointing downward, similar to that shown in \cite{Calvert2021}. 
For adaptation to sugarcane, the AutoWeed detection units were retrofitted onto spray booms centred on the interrow (i.e. in between crop rows). This allowed clear vision of weeds in the row underneath the sugarcane leaf canopy and in the interrow soil area, as shown in Fig.~\ref{fig:classification_approach}. Each dataset collection took approximately one hour, during which a few hundred thousand images (see Table \ref{tab:trial_data}) were collected from several sugarcane rows.

% In the six trials presented in our paper, we focused on three types of weed species: nutgrass, grass, and broadleaf weeds. These weeds were selected based on direct consultation with the participating sugarcane grower, as they are the most prominent in sugarcane paddocks in the Burdekin region of QLD, Australia, where our spray trials took place. As shown in Table \ref{tab:trial_data}, nutgrass has been one of the main weeds targeted in our project. This sedge-like weed requires costly herbicide treatment (usually with Sempra) and shares similar colour and shape features with early-stage plant and ratoon sugarcane, which makes it an interesting target for spot-spraying applications using deep learning among sugarcane crops. Other weeds of interest included two species of grasses (summer grass and crowsfoot) and broadleaf weeds (vines, giant pigweed, and sesbania pea), which we treated in a mung bean rotation crop. It is worth noting that while each paddock may have different weeds, targeting these specific weeds was prioritised due to their higher impact and herbicide costs. Additionally, in our trial, we treated all grass and broadleaf weeds in two respective categories, as there was no need to spot-spray specific species of broadleaf or grass weeds.

In the six trials presented in our paper, we focused on annual grasses, broadleaf weeds and nutgrass. The weeds present in our trials were selected based on direct consultation with the participating sugarcane grower, and were common weeds of sugarcane paddocks in the Burdekin region of QLD, Australia. As shown in Table \ref{tab:trial_data}, nutgrass (Cyperus rotundus) has been one of the main weeds targeted in our project. This sedge can be controlled in a sugarcane crop by applying a selective and costly herbicide treatment (halosulfuron-methyl) and shares similar colour and shape features with early stage sugarcane, which makes it an interesting target for spot-spraying applications using deep learning among sugarcane crops. Other weeds of interest included two grass species: summer grass (Digitaria ciliaris) and crowsfoot (Eleusine indica); and broadleaf weeds: Red convolvulus (Ipomoea hederifolia), giant pigweed (Trianthema portulacastrum) and sesbania pea (Sesbania cannabina), which we treated in a mung bean rotation crop. It is worth noting that while each paddock may have different weeds, targeting these specific weeds was prioritised due to their higher impact and herbicide costs. Additionally, in our trial, we grouped all grass and broadleaf weeds in two respective categories, as there was no need to spot-spray specific species of broadleaf or grass weeds.

\begin{table}[hbt!]
    \begin{center}
        \footnotesize
        \begin{tabularx}{\textwidth}{cccccc}
        \hline
        Trial & Weed(s) & Crop & Images Collected & Images Annotated \\
        \hline\hline
        1 \& 2 & Nutgrass & Ratoon sugarcane & 244,346 & 81,800 \\
        \hline
        3 & Nutgrass & Ratoon sugarcane & 369,744 & 8,638 \\
        \hline
        4 \& 5 & \makecell{Grass and\\broadleaf weeds} & Mung bean & 303,362 & 55,030 \\
        \hline
        6 & Nutgrass & Plant sugarcane & 530,004 & 90,858 \\
        \hline
        Total & & & 1,447,456 & 236,326 \\
        \hline
        \end{tabularx}  
    \end{center}
    \caption{Summary of images collected and annotated for each of the six spray trials.  Grass weeds included summer grass and crowsfoot;  broadleaf weeds included vines, giant pigweed, and sesbania pea.}
    \label{tab:trial_data}
\end{table}

\subsubsection{Dataset annotation for DL training}
%Discuss the labelling approach etc
Following data collection, a group of annotators manually annotated the presence of target weeds in each weed training dataset in a binary fashion, i.e. labelled images with weed present as weed, and the others as non-target. 
A tiled classification approach was chosen whereby collected images were split into 2 $\times$ 2 tiles for annotation (Fig.~\ref{fig:classification_approach}a) and 1 $\times$ 2 tiles for inference during the use of the system for spot-spraying (Fig.~\ref{fig:classification_approach}b). This allowed rapid collection and annotation of high tile counts during dataset collection, and high-speed performance with a batch size of two during inference. It is worth noting that, the exact spatial position of the weeds was not recorded in annotations or used in our spray application. Instead, if an image tile contained a weed, its corresponding spray nozzle was activated, as shown in Fig.~\ref{fig:classification_approach}b. Because the design of the system is made for crops with rows that travel in a straight line direction, this system will accurately identify and spray weeds by only needing to tune the spray on time and spray duration based on the vehicle’s speed of travel.  

\begin{figure}[hbt!]
    \centering
    \includegraphics[width=0.95\textwidth] {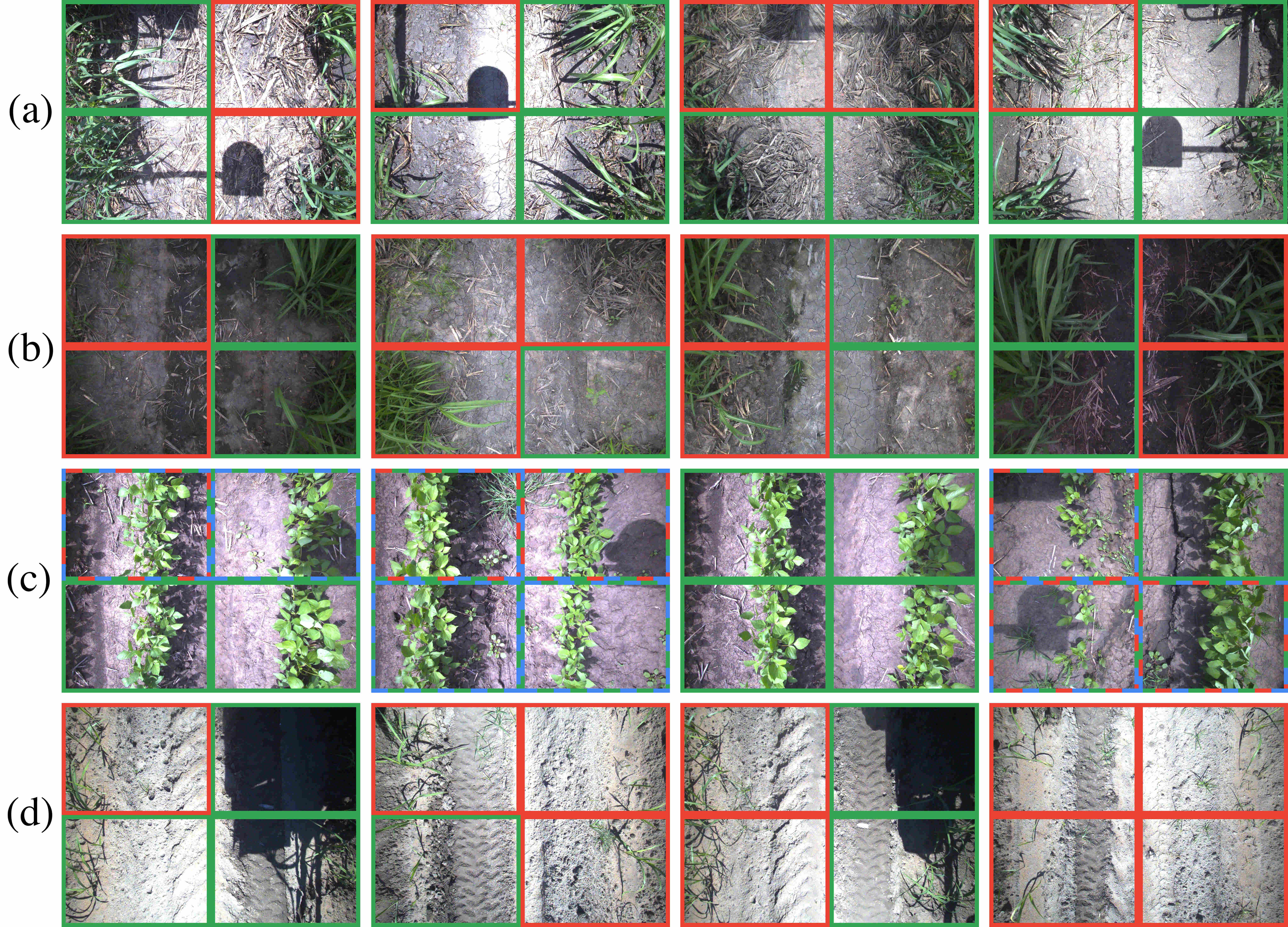} 
    \caption{Sample annotated images from the training datasets for (a) trials 1 \& 2, (b) trial 3, (c) trials 4 \& 5, and (d) trial 6. Red borders indicate a tile contains a target weed and green borders do not. The blue border for (c) indicates a second target weed.\label{fig:data_sample}}
\end{figure}

Annotation was performed using the Computer Vision Annotation Tool (CVAT). This allowed for rapid annotation of approximately 2,000 tile images per hour, and an overall annotation time between 4.5 hours for trial 3 and 45 hours for trial 6. Annotators labelled the presence of target weeds in each tile using a binary approach for spray trials 1, 2, 3 and 6. However, the training dataset for spray trials 4 and 5 required a multi-label classification approach due to the presence of two target weeds (grass and broadleaf) in the dataset. Fig.~\ref{fig:data_sample} provides sample images from the site-specific training datasets that were collected and annotated for each of the trials.

\subsubsection{DL training and validation for weed classification}\label{DLtrain}
Following the labelling process, the training datasets for each trial were randomly split into 80\% and 20\% subsets of training and validation, where 80\% of the labels were set aside for training and 20\% of the labels were set aside for model validation. The class distribution between training and validation subsets was stratified such that the ratio of the weed class(es) to the negative class was (i.e. without weeds) consistent across both subsets.

To allow for benchmarking of accuracy metrics for researchers, machine learning datasets are usually split three ways with training and validation subsets, and a test subset that is heldout to test the model. We have foregone this data splitting approach here, to train models on the most data possible. Furthermore, the test set for this work is the real-time in situ data when evaluating spot-spraying performance in the field.

% \begin{figure}[hbt!]
%     \centering
%     \begin{subfigure}[b]{0.45\textwidth}
%         \centering
%         \includesvg[inkscapelatex=false, width=\textwidth]{./fig7a.svg}
%         \caption{}
%     \end{subfigure}
%     \hfill
%     \begin{subfigure}[b]{0.45\textwidth}
%         \centering
%         \includesvg[inkscapelatex=false, width=\textwidth]{./fig7b.svg}
%         \caption{}
%     \end{subfigure}
%     \\
%     \begin{subfigure}[b]{0.45\textwidth}
%         \centering
%         \includesvg[inkscapelatex=false, width=\textwidth]{./fig7c.svg}
%         \caption{}
%     \end{subfigure}
%     \hfill
%     \begin{subfigure}[b]{0.45\textwidth}
%         \centering
%         \includesvg[inkscapelatex=false, width=\textwidth]{./fig7d.svg}
%         \caption{}
%     \end{subfigure}
%     \caption{Training and validation accuracy versus epoch (left axis) and training and validation loss versus epoch (right axis) during the training process for (a) spray trials 1 and 2, (b) spray trial 3, (c) spray trials 4 and 5 and (d) spray trial 6. \label{fig:training_curves}}
% \end{figure}

The TensorFlow machine learning backend was used together with the Python-based high-level API Keras, to train a MobileNetV2 architecture following the training methodology of \cite{chen2022performance,Calvert2021}. 
MobileNetV2 is a deep learning model designed for efficient image classification tasks, particularly on mobile and embedded devices. It employs an architecture that balances high accuracy with low computational cost, using depthwise separable convolutions and inverted residuals. The loss function commonly used for training MobileNetV2 is categorical cross-entropy, which measures the difference between the predicted probability distribution and the true distribution of the class labels.
On average, the training process took $4-5$ hours on an NVIDIA GTX 1080Ti Graphical Processing Unit (GPU). %The result of the training process is illustrated with time series plots of training and validation metrics versus successive epochs in Fig.~\ref{fig:training_curves}.
Early stopping was used to halt training when the validation loss failed to decrease after 16 successive epochs. %For each trial, the MobileNetV2 model from the epoch with the lowest validation loss was selected for field implementation.

\begin{figure}[hbt!]
    \centering
    \begin{subfigure}[b]{0.45\textwidth}
        \centering
        \includegraphics[width=\textwidth]{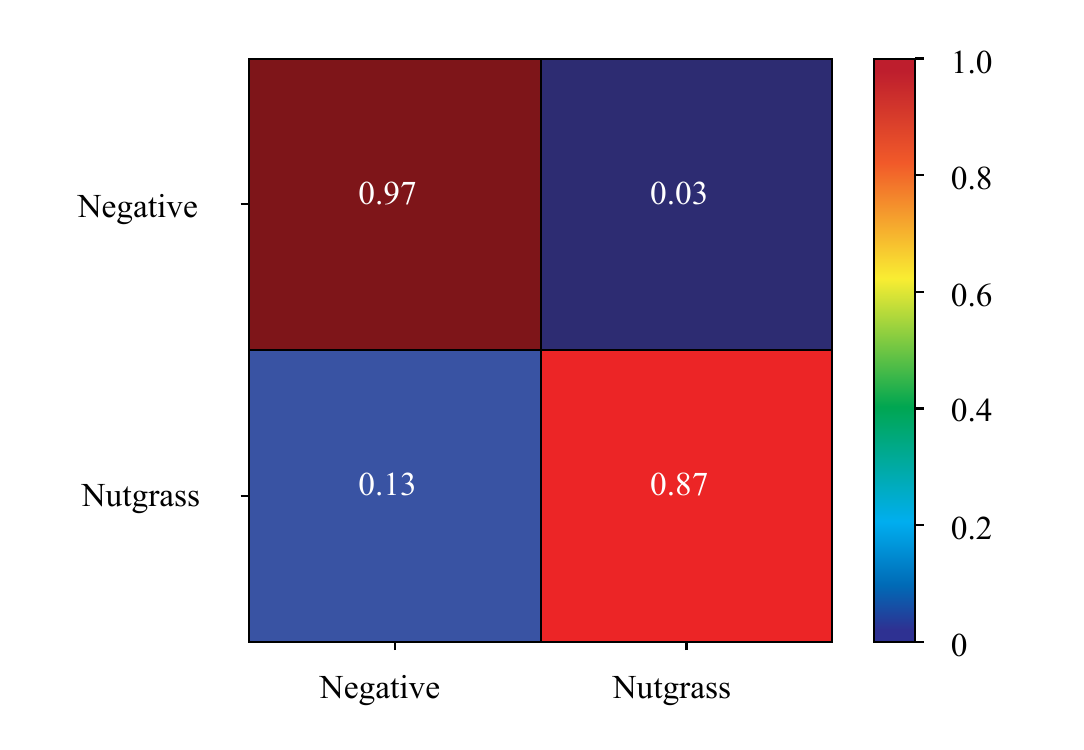}
        \caption{}
    \end{subfigure}
    \hfill
    \begin{subfigure}[b]{0.45\textwidth}
        \centering
        \includegraphics[width=\textwidth]{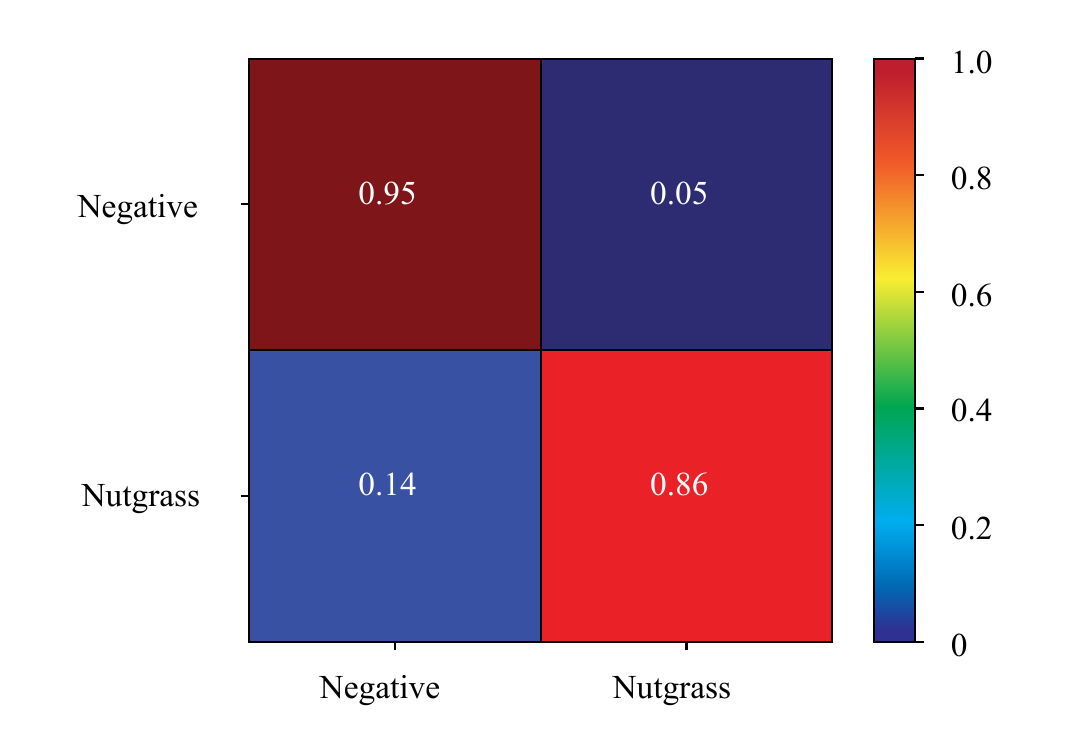}
        \caption{}
    \end{subfigure}
    \\
    \begin{subfigure}[b]{0.45\textwidth}
        \centering
        \includegraphics[width=\textwidth]{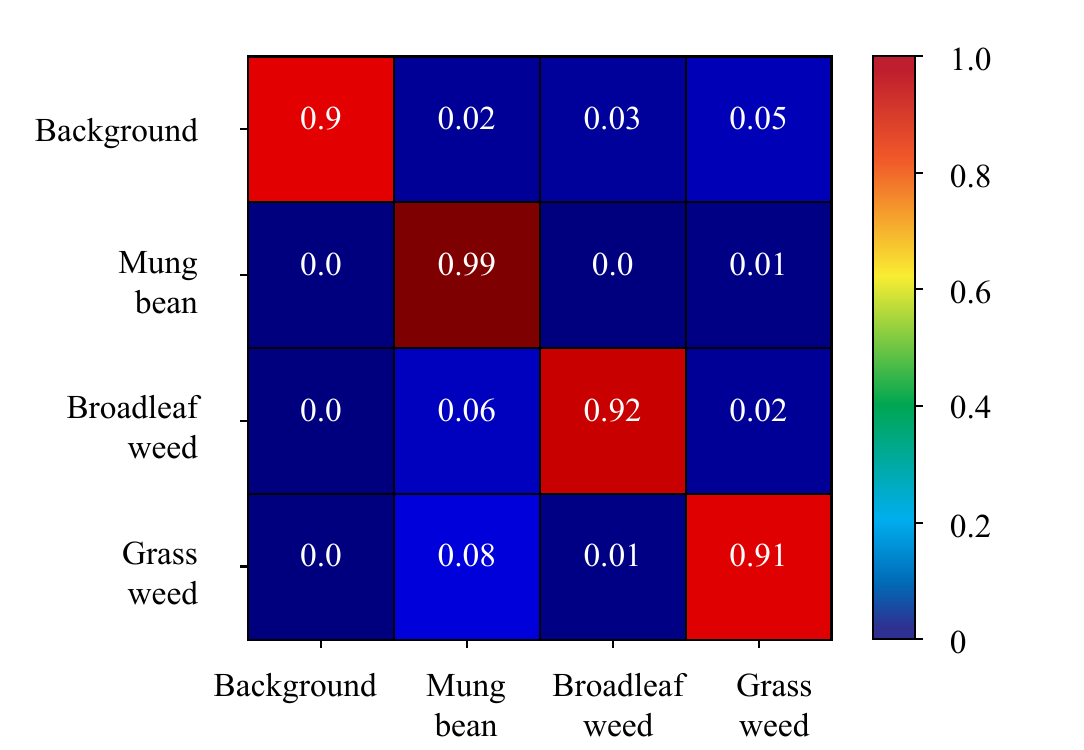}
        \caption{}
    \end{subfigure}
    \hfill
    \begin{subfigure}[b]{0.45\textwidth}
        \centering
        \includegraphics[width=\textwidth]{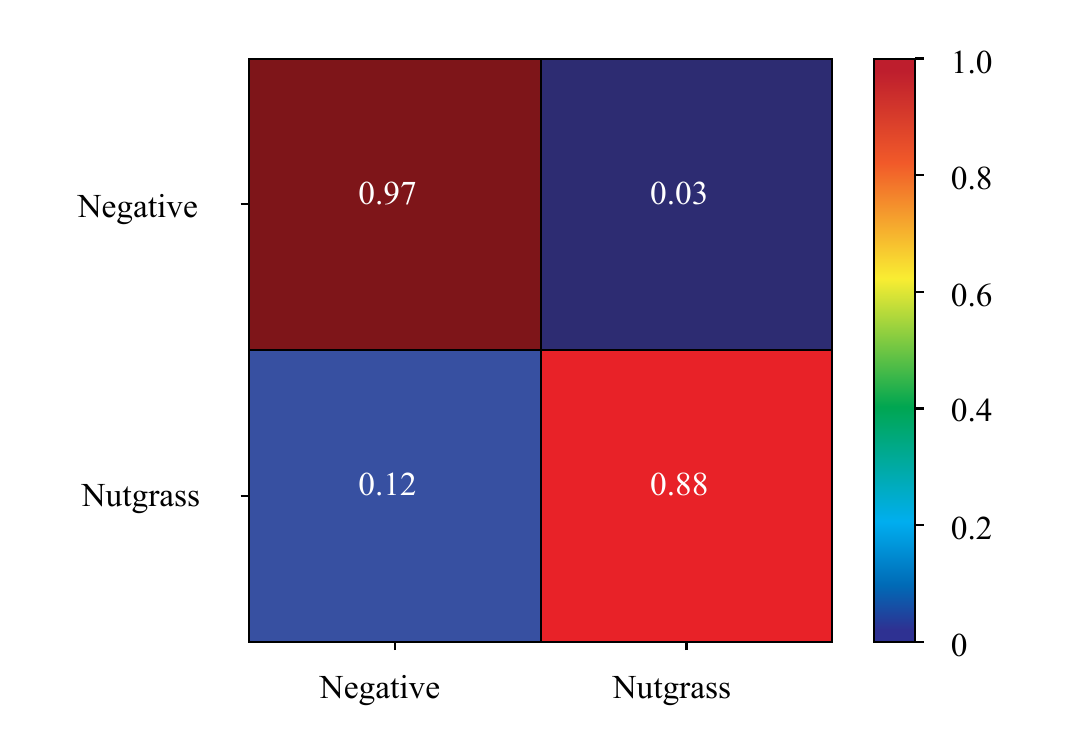}
        \caption{}
    \end{subfigure}
    \caption{ The confusion matrices evaluating the performance of the MobileNetV2 weed detection models in lab inference on 20\% of the labelled validation data. Negative (in a-b, and d) and Background (in c) refer to instances where there is no weed present in the image. The model may incorrectly identify an image with no weed as containing weed (false positive), or miss a weed (false negative).  Here, (a) shows results from spray trials 1 and 2, (b) from spray trial 3, (c) from spray trials 4 and 5, and (d) from spray trial 6. \label{fig:confusion_matrices}}
\end{figure}

Fig.~\ref{fig:confusion_matrices} presents the confusion matrices for each collected dataset, which captures how well each trained model classifies its target weed(s). This data shows lab inference using the MobileNetV2 model from the epoch with the lowest validation loss, which is also used for field implementation. We utilised 20\% of the annotated data to test model performance shown in Fig.~\ref{fig:confusion_matrices}. The data volume used for the various trials can be calculated from the information provided in Table 1. This volume ranges from 1727 images (20\% of 8638) to test the model developed for detecting nutgrass in trial 3, to 18172 images for testing the nutgrass detection model developed for trial 6.

\subsubsection{In-field spot-spraying trials}
For each trial shown in Table \ref{tab:trial_data}, the MobileNetV2 model from the training process with the epoch with the lowest validation loss was selected for field implementation.

The light-weight MobileNetV2 architecture was chosen, similar to other works \cite{chen2022performance,Calvert2021}, because it offers the high-speed inference required to achieve real-time spot spraying while travelling at up to 8 km per hour. When running on an NVIDIA Jetson Nano embedded device, the MobileNetV2 architecture performs inference at $21.9$ ms per image or $45.7$ frames per second.

Our in-field spot-spraying trials demonstrated that a model with high real-world inference accuracy is suitable for spot-spraying target weeds in the field for two main reasons. First, most field variations, such as ambient light, mechanical vibration, and background complexity, are captured in the large training dataset collected for each trial (see Table~\ref{tab:trial_data}). Second, since the deep learning model has multiple opportunities to see the same weed due to the high frame processing rate and capturing images from the same view multiple times, the weed hit rate in the field is even higher than the per-image model accuracy. This is evidenced by the data presented in the next Section, which shows a high weed knockdown hit rate for spot-spraying, consistent with the accuracy of the deep learning models trained on the dataset.

There are also several sources of time delay from the moment that an image is captured to the time that the chemical hits the target weed. These include image acquisition time (i.e. the time it takes for the image to be captured and made available for processing), pre-processing time (i.e. the time it takes to prepare the image format for inference on the embedded GPU device), inference time (i.e. the time to perform inference using the embedded GPU target device and return a result indicating the presence of a weed or not) and solenoid response time (i.e. the time to send a command to engage a solenoid and have the solenoid electrically engaged).

For this work, we have measured the average time for each of these delays as shown in Table \ref{tab:time_data}. The total average response time from image capture to weed spray is 58.16 ms, using the MobileNetV2 architecture. While travelling at 8 km/hr, the spray vehicle would have only moved approximately 129.2mm over this response time. To ensure adequate coverage of the target weed, we activate the sprayer as quickly as possible to ensure that the sprayer is fully on before the centre of the sprayer passes over the identified weed target. We also instituted a spray duration, that is based on the vehicle speed, so that the sprayer remains on until the sprayer has passed over the target weed to give it full coverage of herbicide. Typically, a spray duration of 0.45 seconds was used for a travel speed of 8 km/hr. This ensured spray sections of approximately 1 metre for identified targets.

\begin{table}[hbt!]

    \begin{center}
        \footnotesize
        \begin{tabularx}{10cm}{lcc}
        \hline
        Time Measurement & Average (ms) & Std. Dev. (ms) \\
        \hline\hline
        Image acquisition time & 5.85 & 0.75 \\
        \hline
        Image pre-processing time & 8.88 & 0.05 \\
        \hline
        Inference time & 21.90 & 5.53 \\
        \hline
        Solenoid response time & 21.53 & 1.70 \\
        \hline
        Total execution time & 58.16 & 5.83 \\
        \hline
        \end{tabularx}  
    \end{center}
    \caption{ Average time measurements for the spot spraying workflow from image acquisition to solenoid activation showing the average and standard deviation measurements in milliseconds.}
    \label{tab:time_data}
\end{table}

\subsection{Weed knockdown efficacy and herbicide usage analyses}
Participating growers for the trial work were recruited by Sugar Research Australia, whose active presence in the Burdekin region garnered great interest in the project from local growers. Trial paddocks were restricted to early-stage ratoon sugarcane, plant sugarcane, or rotational crops with the presence of common and priority weeds in the Burdekin region. The aforementioned six experimental trials were conducted with details provided in Table~\ref{tab:trials}.

\begin{table}
    \begin{center}
        \tiny
        \begin{tabularx}{\textwidth}{ccccccccc}
        \hline
        Trial & Date & Weed & Crop & Herbicide & \makecell{Nozzle\\ type} & \makecell{Spray\\ boom\\ width} & \makecell{Blanket\\ spraying\\ runs} & \makecell{Spot\\ spraying\\ runs} \\
        \hline\hline
        1 & 11/11/2021 & Nutgrass & \makecell{Ratoon\\ sugarcane} & Sempra & Irvin leg & 13-row & $2\times1.25\ \unit{ha}$ & $2\times1.25\ \unit{ha}$ \\
        \hline 
        2 & 11/11/2021 & Nutgrass & \makecell{Ratoon\\ sugarcane} & Sempra & Irvin leg & 4-row & $2\times1.25\ \unit{ha}$ & $2\times1.25\ \unit{ha}$ \\
        \hline 
        3 & 11/12/2021 & Nutgrass & \makecell{Ratoon\\ sugarcane} & Krismat & Irvin leg & 4-row & $2\times1.10\ \unit{ha}$ & $2\times1.10\ \unit{ha}$ \\
        \hline 
        4 & 21/3/2022 & \makecell{Grass\\ weeds} & Mung bean & Verdict & \makecell{Flat\\ boom} & 4-row & $2\times0.9\ \unit{ha}$ & $2\times0.9\ \unit{ha}$ \\
        \hline 
        5 & 21/3/2022 & \makecell{Broadleaf\\ weeds} & Mung bean & Blazer & \makecell{Flat\\ boom} & 4-row & $2\times0.9\ \unit{ha}$ & $2\times0.9\ \unit{ha}$ \\
        \hline
        6 & 2/8/2022 & Nutgrass & \makecell{Plant\\ sugarcane} & Sempra & \makecell{Flat\\ boom} & 4-row & $2\times0.75\ \unit{ha}$ & $2\times0.75\ \unit{ha}$  \\
        \hline
        \end{tabularx}  
    \end{center}
    \caption{Summary of the experimental trials that were completed in this work.}

    \label{tab:trials}
\end{table}

\begin{figure}
	\centering
	\includegraphics[width=0.8\textwidth] {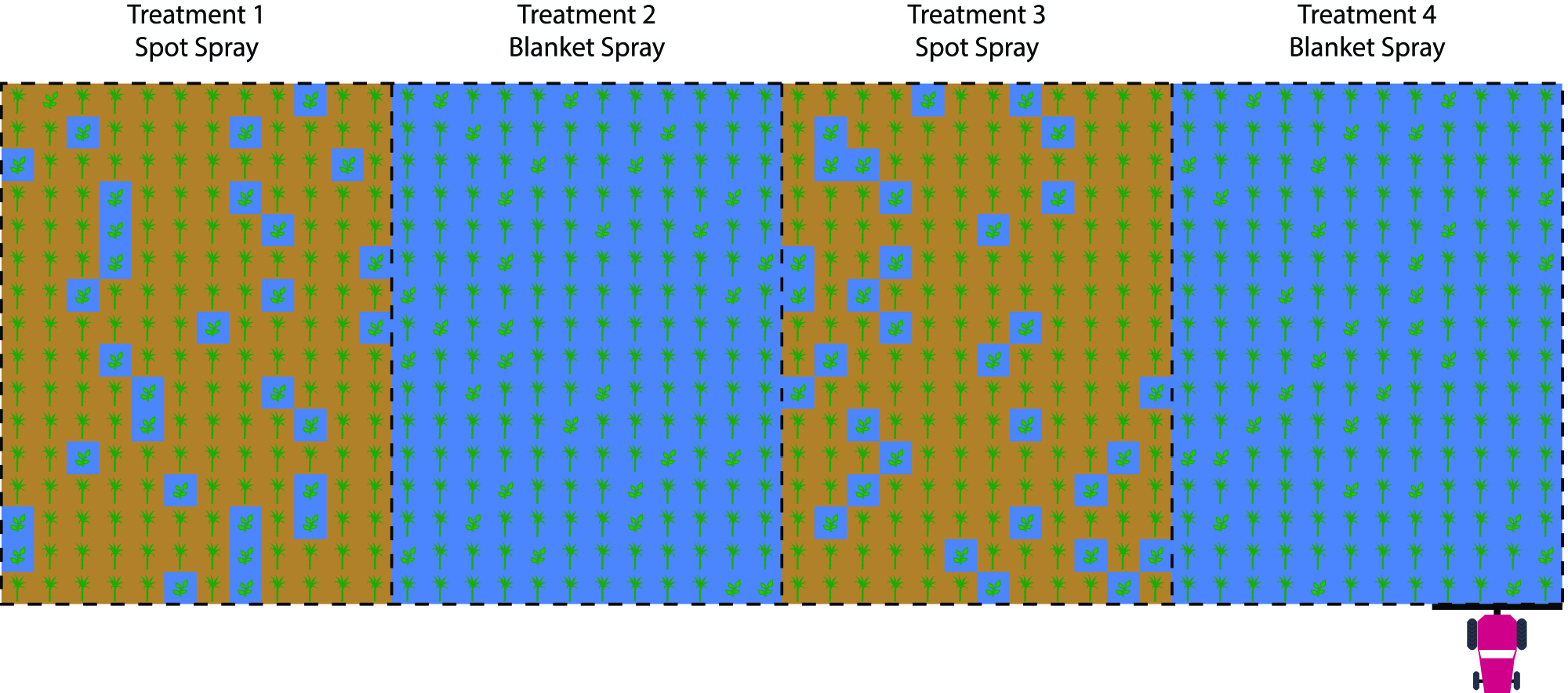}
	\caption{An illustration of the replicated-strip trial design which was implemented for all spray trials with four 12-13 row alternating treatments of spot spraying and blanket spraying.}
	\label{fig:trial_design}
\end{figure}

For each trial, a 12-13 row  replicated strip trial design was implemented as illustrated in Fig.~\ref{fig:trial_design}.  For each 12-13 row strip, chemical treatments were alternated between spot spraying and blanket spraying while using the same herbicide mixtures. Trials 1 and 2 were conducted using a 13-row self-propelled sprayer, while the other trials utilised a 4-row tractor boom sprayer. Trials 4 and 5 were conducted in the morning and afternoon, respectively, on the same crop on the same day. In this situation, two different chemicals were required to control different types of weeds.

Herbicide usage was measured for each treatment using the flow rate controller and tank measurements on the retrofitted spray booms. Custom-made electronics on-board the vehicle also recorded the GPS location and duration of each nozzle activation which allowed visualisation of spray application maps after each trial.

\begin{figure}[hbt!]
    \centering
    \begin{subfigure}[b]{0.243\textwidth}
        \centering
        \includegraphics[width=\textwidth]{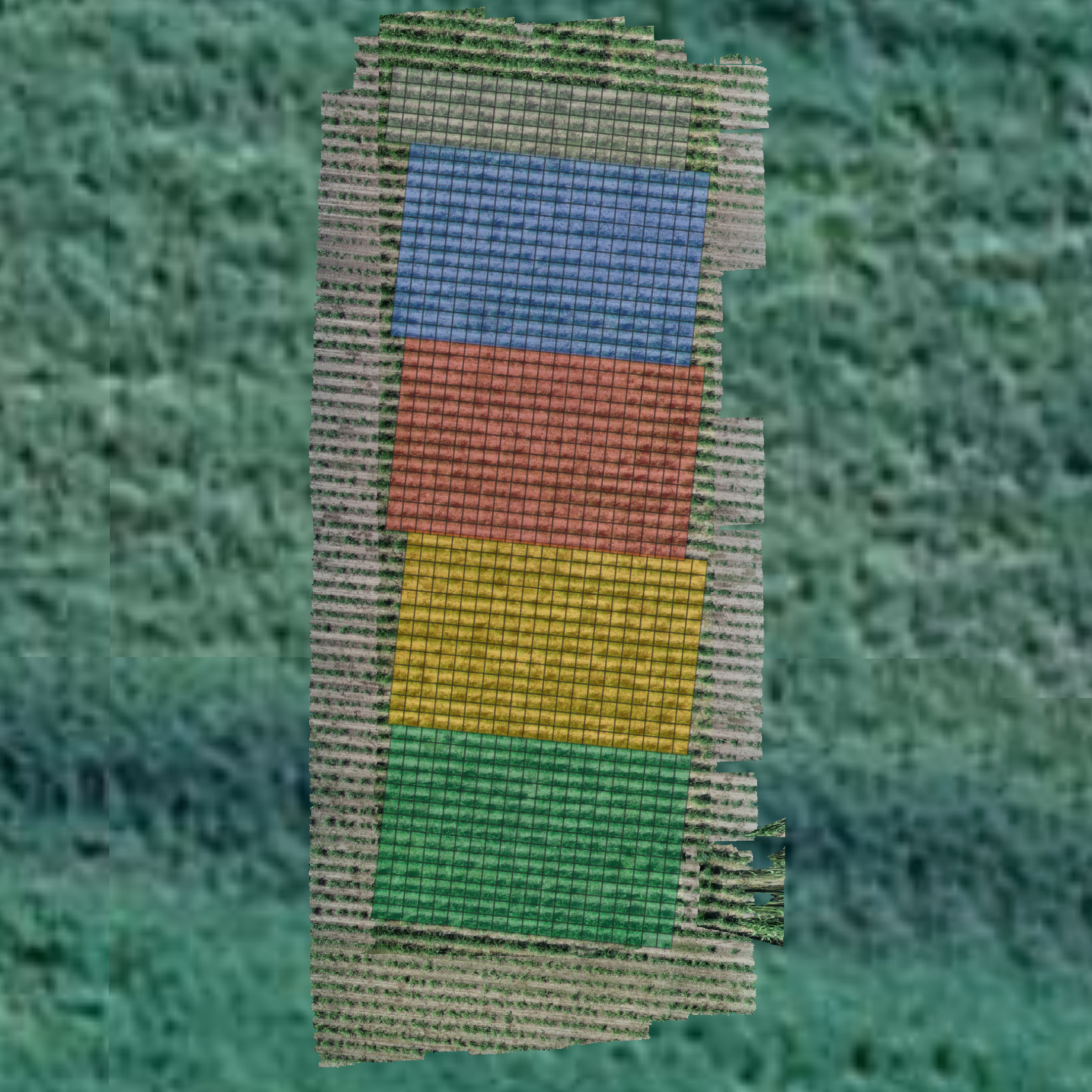}
        \caption{}
    \end{subfigure}
    \hfill
    \begin{subfigure}[b]{0.243\textwidth}
        \centering
        \includegraphics[width=\textwidth]{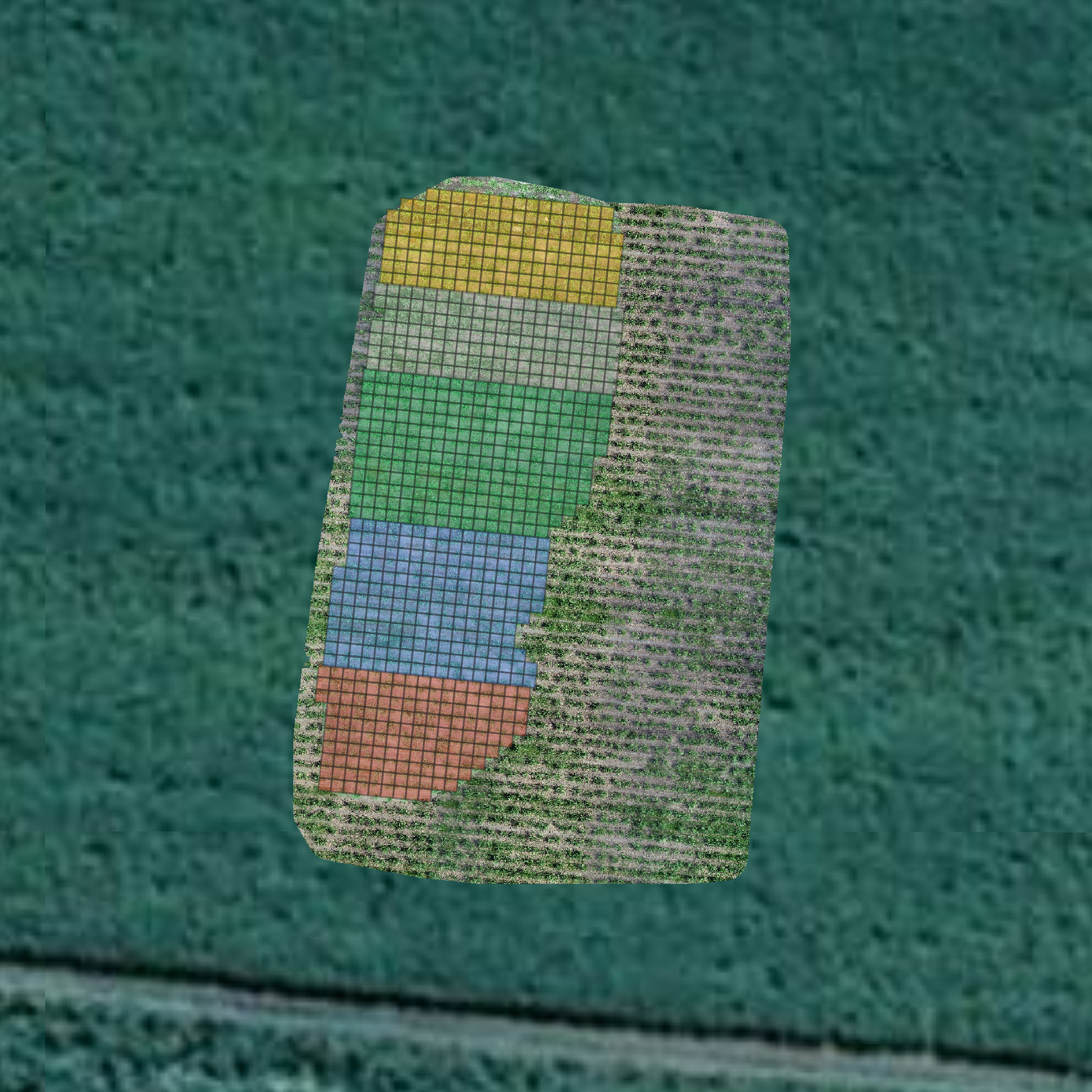}
        \caption{}
    \end{subfigure}
    \hfill
    \begin{subfigure}[b]{0.243\textwidth}
        \centering
        \includegraphics[width=\textwidth]{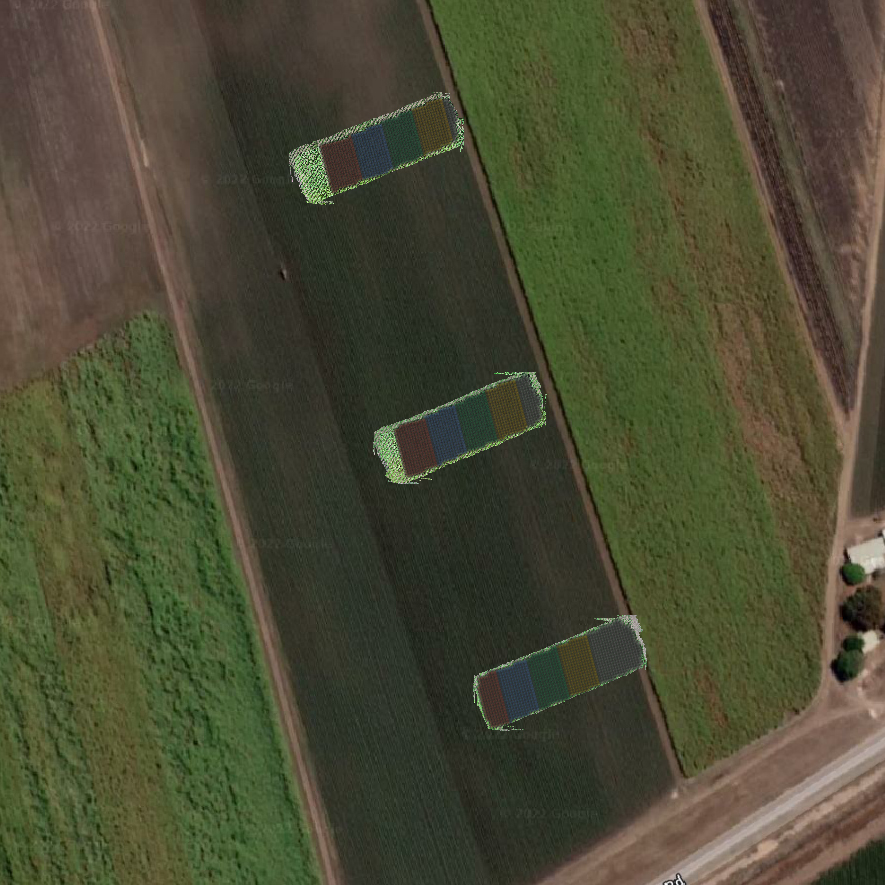}
        \caption{}
    \end{subfigure}
    \hfill
    \begin{subfigure}[b]{0.243\textwidth}
        \centering
        \includegraphics[width=\textwidth]{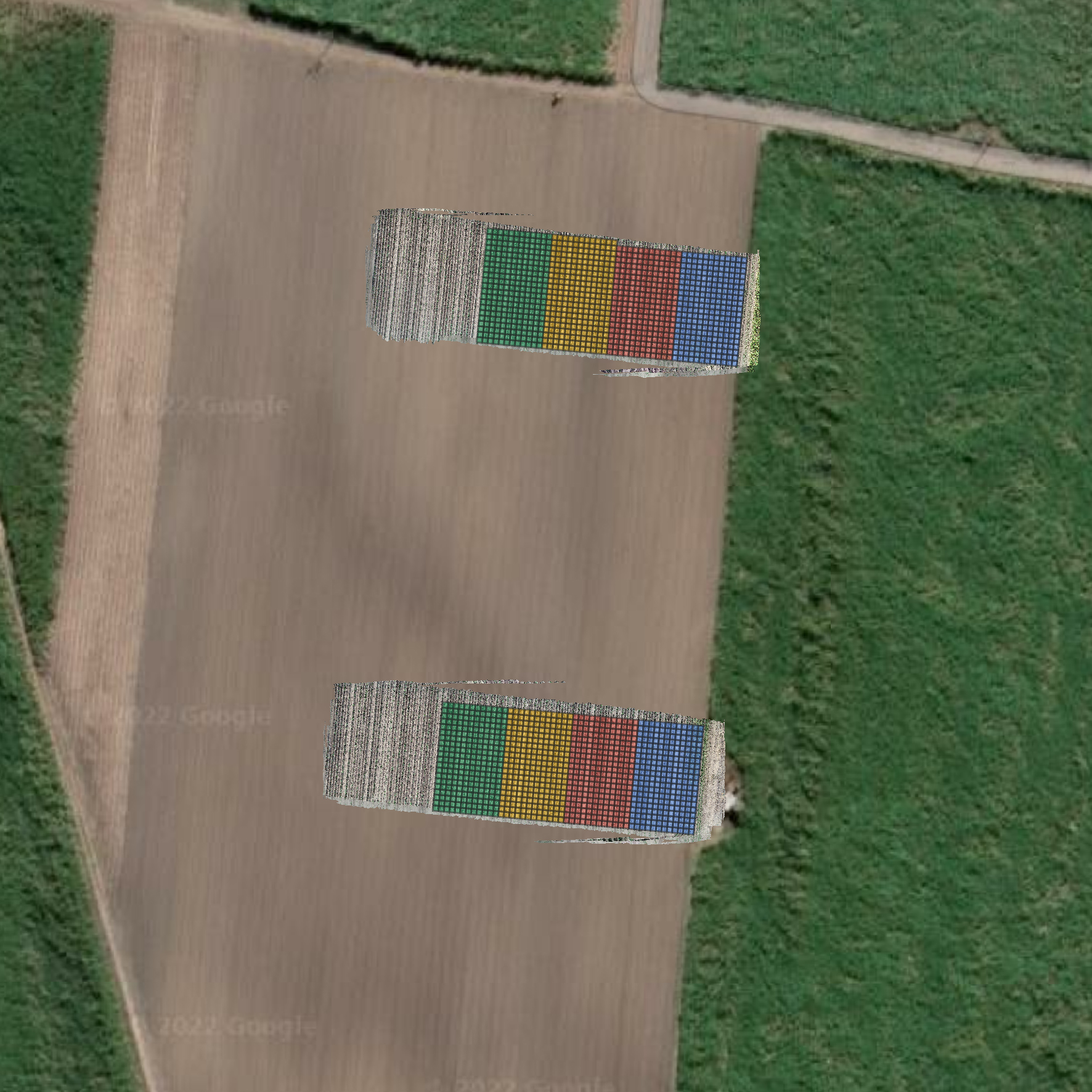}
        \caption{}
    \end{subfigure}
    \caption{The before and after UAV maps for (a) spray trial 2, (b) spray trial 3, (c) spray trials 4 and 5 and (d) spray trial 6. The different colours indicate different treatment areas. Maps for trial 1 were not obtained due to a technical issue with the UAV.\label{fig:drone_maps}}
\end{figure}

To quantify the efficacy of each treatment on weeds,  UAVs were deployed to capture high-resolution imagery from random locations along the trial paddock across all treatments before and after spraying. ``Before maps'' were collected on the day of the trial before applying the treatments. ``After maps'' were collected between 6-14 days after the treatment depending on the time it took visual symptoms to manifest on the targeted weeds after herbicide application. Fig.~\ref{fig:drone_maps} presents the collected drone maps from each target trial site as part of the weed knockdown efficacy analysis. OpenDroneMap \cite{odm2020} was used to generate orthomosaics from the raw drone imagery and QGIS \cite{qgis2022} was used to georeference the before and after orthomosaics and extract comparative $1.5 \times 1.5$ metre $-$ $1.6 \times 1.6$ metre images from the maps. A team of annotators were then able to manually annotate each image counting the number of target weeds sprayed and target weeds missed. The weed knockdown hit rate for each treatment was calculated as
\begin{equation}\label{eq:hitrate}
\mathrm{hit \: rate} = \frac{\mathrm{weeds_{sprayed}}}{\mathrm{weeds_{sprayed}} + \mathrm{weeds_{missed}}}. 
\end{equation}
The weed knockdown efficacy of spot spraying compared to blanket spraying was calculated as
\begin{equation}\label{eq:knockdown}
\mathrm{efficacy} = \frac{\mathrm{hit \: rate_{spot \: spray}}}{\mathrm{hit \: rate_{blanket \: spray}}}.
\end{equation}

\subsection{Water quality measurements}

The impact of the spot spraying on water quality was measured by sampling the run-off water from the first irrigation event that followed the spraying. Unfortunately, irrigation water sampling could not be performed for trials 1 -- 2 due to unexpected rain that generated runoff that could not be captured. Therefore, water quality data are only available for trials 3 -- 6.

The trial farms use a furrow irrigation method, whereby water is released at the top of each row and allowed to run along the ground, gradually flowing downhill under the influence of gravity. Each farmers' normal infrastructure was used for the irrigation. Irrigation was timed to occur as soon as possible after the spraying while complying with each herbicide's label restrictions. This was done to maximise the concentration of herbicide in the sampled water to increase the likelihood of the concentration being sufficiently above the limit of detection to enable an accurate measurement. Therefore, the runoff measurements represent the worst case scenario for water quality. However, all irrigation timing complied with the label restrictions and therefore may be used in practice by farmers.

\begin{figure}[hbt!]
    \centering
    \begin{subfigure}[b]{0.4\textwidth}
        \centering
        \includegraphics[width=\textwidth]{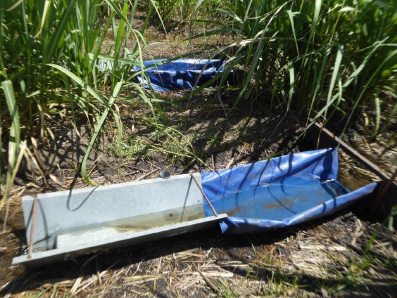}
        \caption{}
    \end{subfigure}
    \hfill
    \begin{subfigure}[b]{0.4\textwidth}
        \centering
        \includegraphics[width=\textwidth]{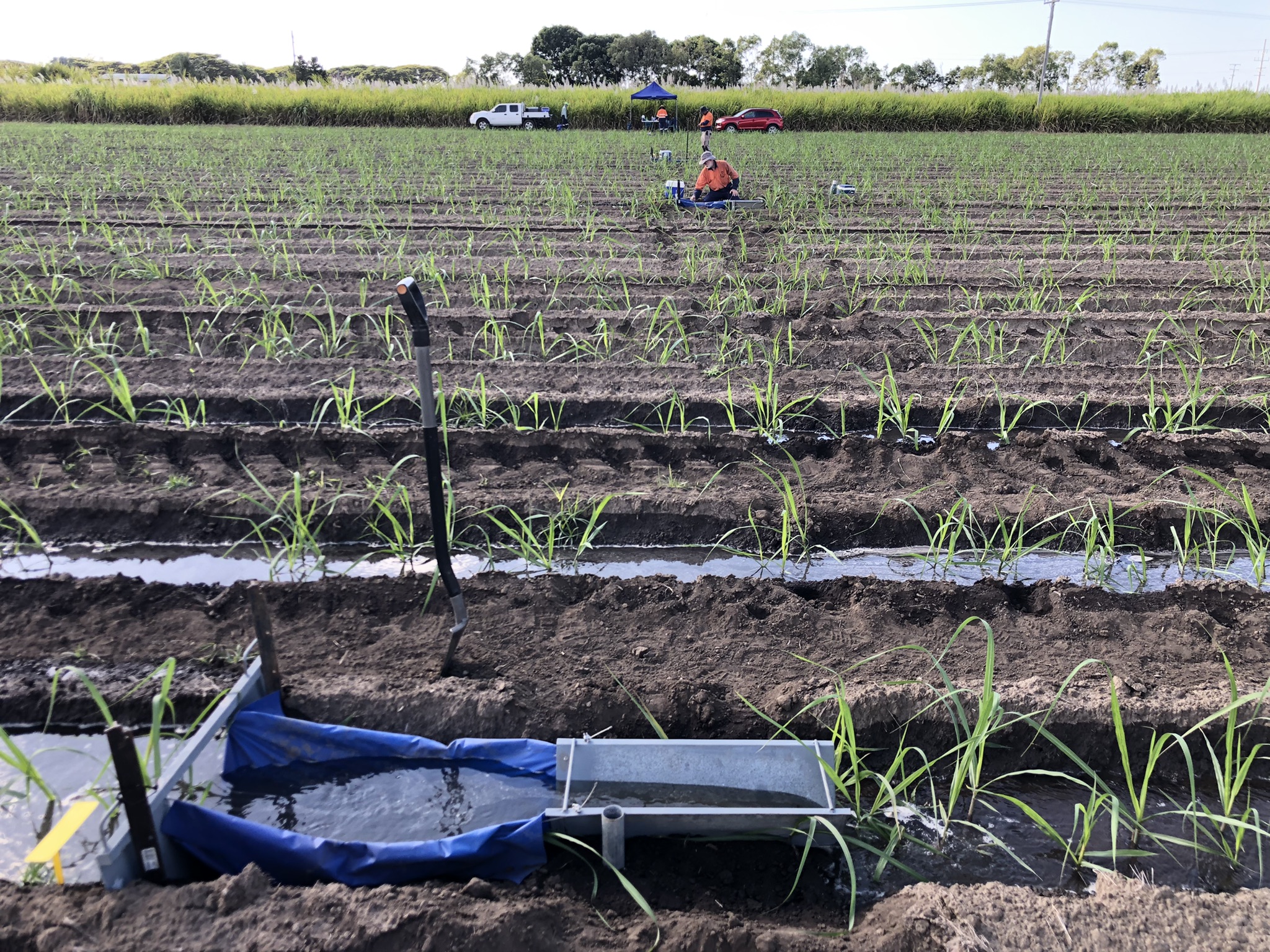}
        \caption{}
    \end{subfigure}
    \caption{(a) RBC flumes installed by SRA staff between sugarcane rows to allow sampling of water quality for each treatment. (b) In-field water quality sampling taking place after the first irrigation event following spray trial 6. \label{fig:flumes}}
\end{figure}

To measure the flow rate and collect water samples, RBC (Replogle, Bos, and Clemmens) flumes were installed on selected rows per plot as shown in Fig.~\ref{fig:flumes}. 
% Care was taken not to re-enter the sprayed paddock until the end of the exclusion period as specified on the label for each chemical. 
The selected rows were typically close to the centre of each trial strip, to ensure minimal cross-contamination from rows belonging to another treatment.

The start time of the irrigation and the flow rate were decided in consultation with each grower so that the water would reach each flume when the staff were ready to collect the samples. Once water reached the flume, the water height within it was monitored every five minutes to calculate the flow rate. The first water sample was taken 1 minute after the runoff began, and further collections occurred on a flow basis (i.e. after a specified volume of water had passed the flume, to capture a total of 1 to 3 L composite sample throughout the runoff event). For each sample, 100 mL of runoff water was collected in a glass container. Each 100 mL sample was mixed into a 1 L glass bottle to create a composite sample, with fresh 1 L bottles used if needed, until the end of the runoff event. The total event duration was approximately 5 hours for trial 3 and 2-3 hours for trials 4 to 6. The water bottles were kept cold using an ice block in a portable insulated cooler box and wrapped with aluminium foil to keep the liquid away from sunlight.

Water samples were refrigerated at 4-6 \textdegree C, and a representative subsample was dispatched by overnight road freight (while being kept cold) to the Sugar Research Australia laboratories for chemical analysis to determine the concentration of each herbicide's active ingredients. After 0.45 \textmu m filtration, concentration was measured using High-Performance Liquid Chromatography (HLPC) on a Shimadzu Nexera X2 and LCMS-2020 system with a Kinetex 1.7 \textmu m C18, 100 x 2.1 mm LC column. The analyte was eluted using an ultra-high pressure gradient method with a clean-up step over a period of 30 minutes using mobile phases consisting of 0.2\% formic acid in water (A), and 0.2\% formic acid in (5:95 v/v) water and acetonitrile mixture (B) from 8 to 95\% (B) at a flow rate of 0.200 mL/min at 40°C.

Analyte detection was performed by a single quadrupole mass spectrometer with a Dual Ion Source (DUIS) probe in both positive and negative Selective Ion Monitoring (SIM) modes using the LabSolution software. Quantitation was achieved by internal standardisation using stable-labelled isotope with a calibration range between 0.0005 to 0.200 mg/L.
               
\section{Results} \label{Sec:Results}

\begin{figure}[hbt!]
    \centering
    \includegraphics[width=\textwidth] {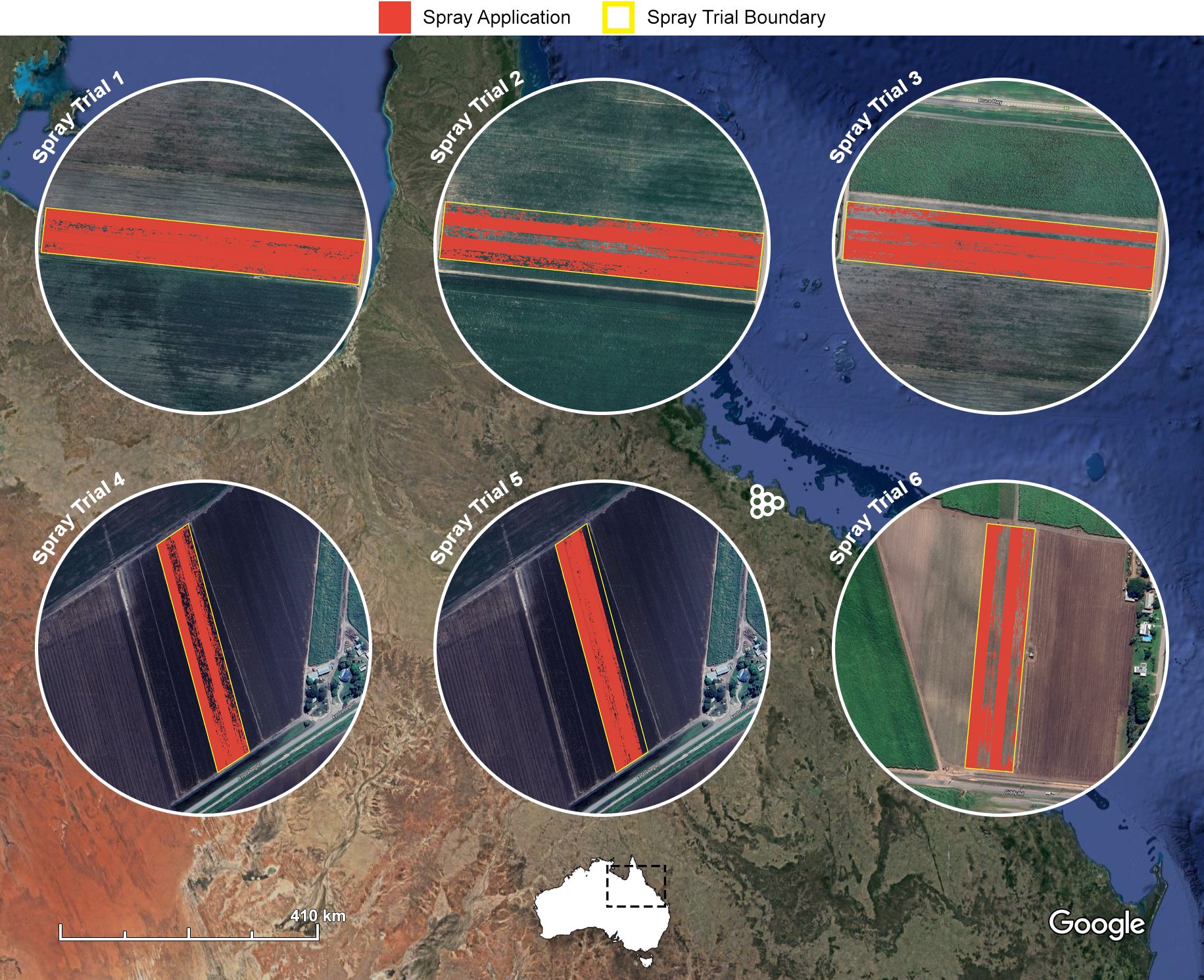}
    \caption{Spray application maps for each of the six spray trials which were spread across the Burdekin region in Queensland Australia. Gaps in the spray application indicate regions without weeds, and hence opportunities to reduce herbicide use. \label{fig:spray_maps}}
\end{figure}

Herbicide application during the six field trails is visually summarised in Fig.~\ref{fig:spray_maps}. The maps show the application of herbicide across each of the sites, and the alternating strips of blanket and spot spraying are visible in the images. 

Table~\ref{tab:efficacy_results} gives the weed knockdown hit rate for each spraying method (based on Eq. \ref{eq:hitrate}) and spot spraying knockdown efficacy (based on Eq. \ref{eq:knockdown}) from the trial work. It also shows a significant reduction in herbicide usage using spot spraying compared to blanket spraying.  Table~\ref{tab:wq_results} summarises the water quality improvement results from these trials.

\begin{sidewaystable}
    \footnotesize
    \begin{center}
        \begin{tabularx}{\textwidth}{ccccccccc}
        \hline
        Trial & \makecell{Weed /\\Crop} & Herbicide & \makecell{Blanket\\ spraying\\ knockdown\\ hit rate (\%)} & \makecell{Spot\\ spraying\\ knockdown\\ hit rate (\%)} & \makecell{Blanket\\ spraying\\ herbicide\\ usage (L/ha)} & \makecell{Spot\\ spraying\\ herbicide\\ usage (L/ha)} & \makecell{ Spot\\ spraying\\knockdown\\ efficacy\\ (\%)} & \makecell{Spot\\ spraying\\herbicide\\ reduction\\ (\%)} \\
        \hline\hline
        1$^{*}$ & \makecell{Nutgrass /\\Ratoon\\ sugarcane} & Sempra & - & - & 200 & 177 & - & 11 \\
        \hline 
        2 & \makecell{Nutgrass /\\Ratoon\\ sugarcane} & Sempra & 97 & 95 & 198 & 81 & 98 & 59 \\
        \hline 
        3 & \makecell{Nutgrass /\\Ratoon\\ sugarcane} & Krismat & 97 & 89 & 199 & 183 & 92 & 8 \\
        \hline 
        4 & \makecell{Grass\\ weeds /\\Mung bean} & Verdict & 99 & 96 & 211 & 100 & 97 & 53 \\
        \hline 
        5 & \makecell{Broadleaf\\ weeds /\\Mung bean} & Blazer & 100 & 100 & 211 & 178 & 100 & 16 \\
        \hline
        6 & \makecell{Nutgrass /\\Plant\\ sugarcane} & Sempra & 100 & 96 & 207 & 73 & 96 & 65 \\
        \hline
        Average & Various & Various & 99 & 95 & 204 & 132 & 97 & 35 \\
        \end{tabularx}  
    \end{center}
    \caption{Summary of the spot spraying knockdown efficacy (Eq. \ref{eq:knockdown}) and herbicide usage reduction compared to blanket spraying results for the trials. $^{*}$Note that no knockdown comparison was possible for all spray trial 1 plots due to technical error during the follow up UAV flights.}
    \label{tab:efficacy_results}
\end{sidewaystable}

\begin{sidewaystable}
    \footnotesize
    \begin{center}
        \begin{tabularx}{\textwidth}{lllllllll}
        \hline
        Trial & \makecell{Weed /\\Crop} & \makecell{Herbicide \\commercial \\name / \\active \\ingredient} & \makecell{Mean\\concentration\\in runoff\\for blanket\\spraying\\ (\textmu g/L)} & \makecell{Loads in\\runoff\\for blanket\\ spraying\\ (g/ha)} & \makecell{Mean\\concentration\\ in runoff\\for spot\\ spraying\\ (\textmu g/L)} & \makecell{Loads in\\runoff\\for spot\\ spraying\\ (g/ha)} & \makecell{Reduction\\of mean\\concentration\\in runoff (\%)} & \makecell{Reduction\\of loads in\\ runoff (\%)} \\
        \hline\hline
        1$^{*}$ & \makecell{Nutgrass /\\Ratoon\\ sugarcane} & \makecell{Sempra /\\Halosulfuron} & - & - & - & - & - & - \\
        \hline 
        2$^{*}$ & \makecell{Nutgrass /\\Ratoon\\ sugarcane} & \makecell{Sempra /\\Halosulfuron} & - & - & - & - & - & - \\
        \hline 
        3$^{\dagger}$& \makecell{Nutgrass /\\Ratoon\\ sugarcane} & \makecell{Krismat /\\Ametryn} & 111.5 & 15 & 54.25 & 5.94 & 51\% & 60\% \\
        \hline
        3$^{\dagger}$ & \makecell{Nutgrass /\\Ratoon\\ sugarcane} & \makecell{Krismat /\\Trifloxysulfuron} & 3.56 & 0.48 & 2.13 & 0.23 & 40\% & 51\% \\
        \hline 
        4 & \makecell{Grass\\ weeds /\\Mung bean} & \makecell{Verdict /\\Haloxyfop} & 0.5 & 0.03 & 0.27 & 0.01 & 46\% & 67\% \\
        \hline 
        5 & \makecell{Broadleaf\\ weeds /\\Mung bean} & \makecell{Blazer /\\Acifluorfen} & 28.95 & 2.04 & 21.68 & 0.91 & 25\% & 55\% \\
        \hline
        6 & \makecell{Nutgrass /\\Plant\\ sugarcane} & \makecell{Sempra /\\Halosulfuron} & 0.74 & 0.74 & 0.49 & 0.49 & 34\% & 34\% \\
        \hline
        Average & Various & Various & 29.05 & 3.66 & 15.76 & 1.52 & 39\% & 54\% \\
        \end{tabularx} 
    \end{center}
    \caption{Summary of the water quality results for the trials. $^{*}$Water quality analyses could not be completed for spray trials 1 and 2 due to more than 90mm of unforecasted rain in the days after spraying and before the planned date for water quality sampling. $^{\dagger}$Spray trial 3 used Kristmat which has two active chemical agents present for the water quality analysis.}
    \label{tab:wq_results}
\end{sidewaystable}

\newpage

\newpage

\subsection{Weed knockdown hit rate and efficacy of spot compared to blanket spraying}
\begin{figure}[hbt!]
    \centering
    \includegraphics[width=\textwidth] {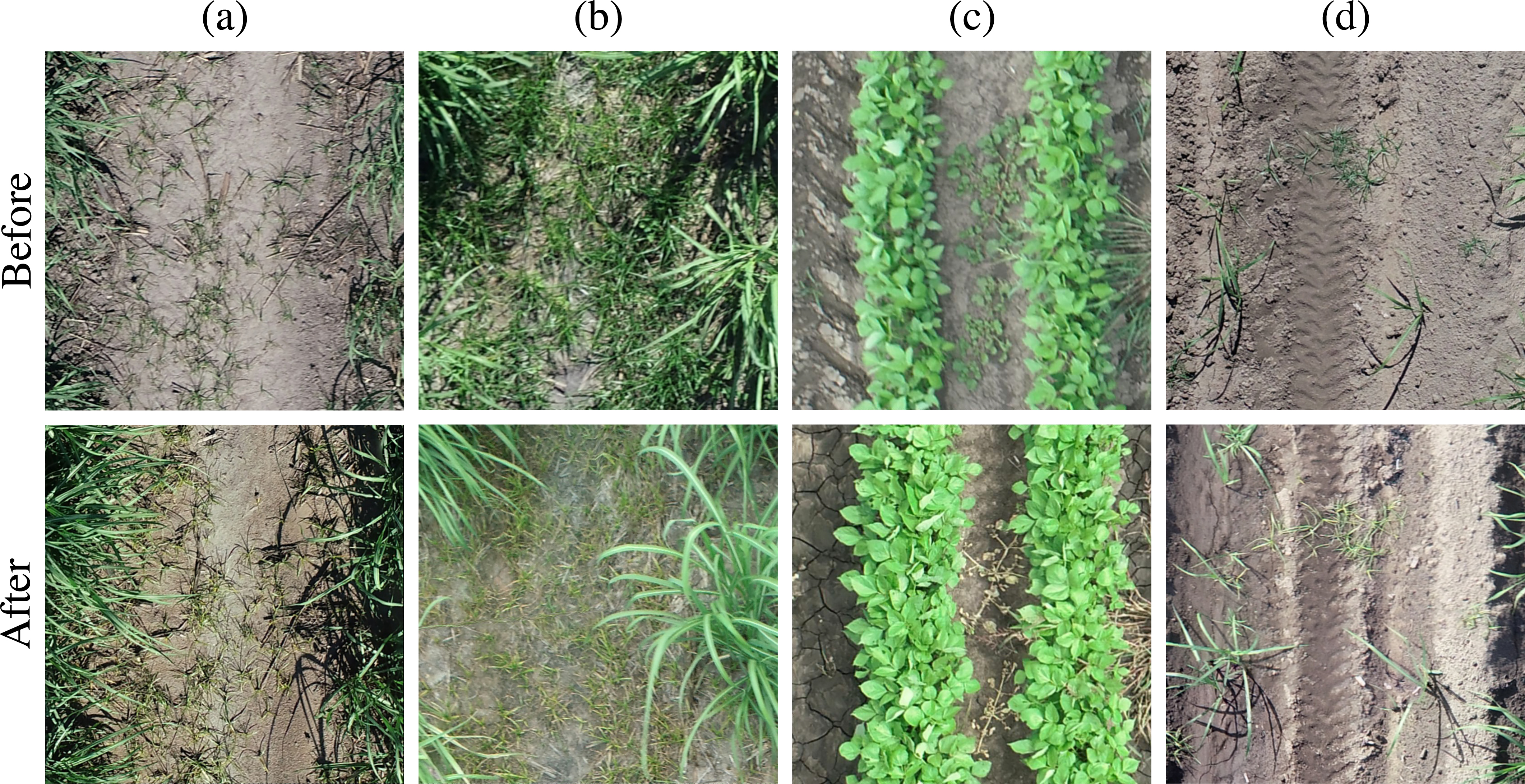}
    \caption{Example UAV images from before (top) and after (bottom) spot spraying for (a) spray trial 2, (b) spray trial 3, (c) spray trials 4 and 5 and (d) spray trial 6 showing the knockdown of target weeds. Note that there was no drone data for trial 1 due to technical issues and trials 4 and 5 use the same drone imagery.\label{fig:knockdown_examples}}
\end{figure}

Weed knockdown hit rate and efficacy were respectively measured using equations \ref{eq:hitrate} and \ref{eq:knockdown} for trials 2 -- 6 using the aforementioned UAV-based approach.  Fig.~\ref{fig:knockdown_examples} displays examples before and after UAV imagery collected from the trials. Weed knockdown could not be measured with this approach for spray trial 1 due to technical issues during the drone flight following the trial. Weed knockdown was visually determined by looking for visible effects of the herbicide on the target weed. The UAV imagery was collected with approximately 2 mm per pixel resolution. Different herbicides generated different visual symptoms on the target weeds at the time of the follow-up UAV mapping. For example, Sempra in trials 2 and 6 is a slow-acting herbicide, and at the time of assessment, symptoms were observed as a slight yellowing of nutgrass. On the other hand, Krismat is a faster acting herbicide, which generated strong yellowing and wilting of nutgrass at the time of assessment. For trials 4 and 5, Verdict led to the destruction of grass weed growth points, and Blazer generated necrotic symptoms on broadleaf weeds.

\begin{figure}[hbt!]
    \centering
    \includegraphics[width=0.98\textwidth]{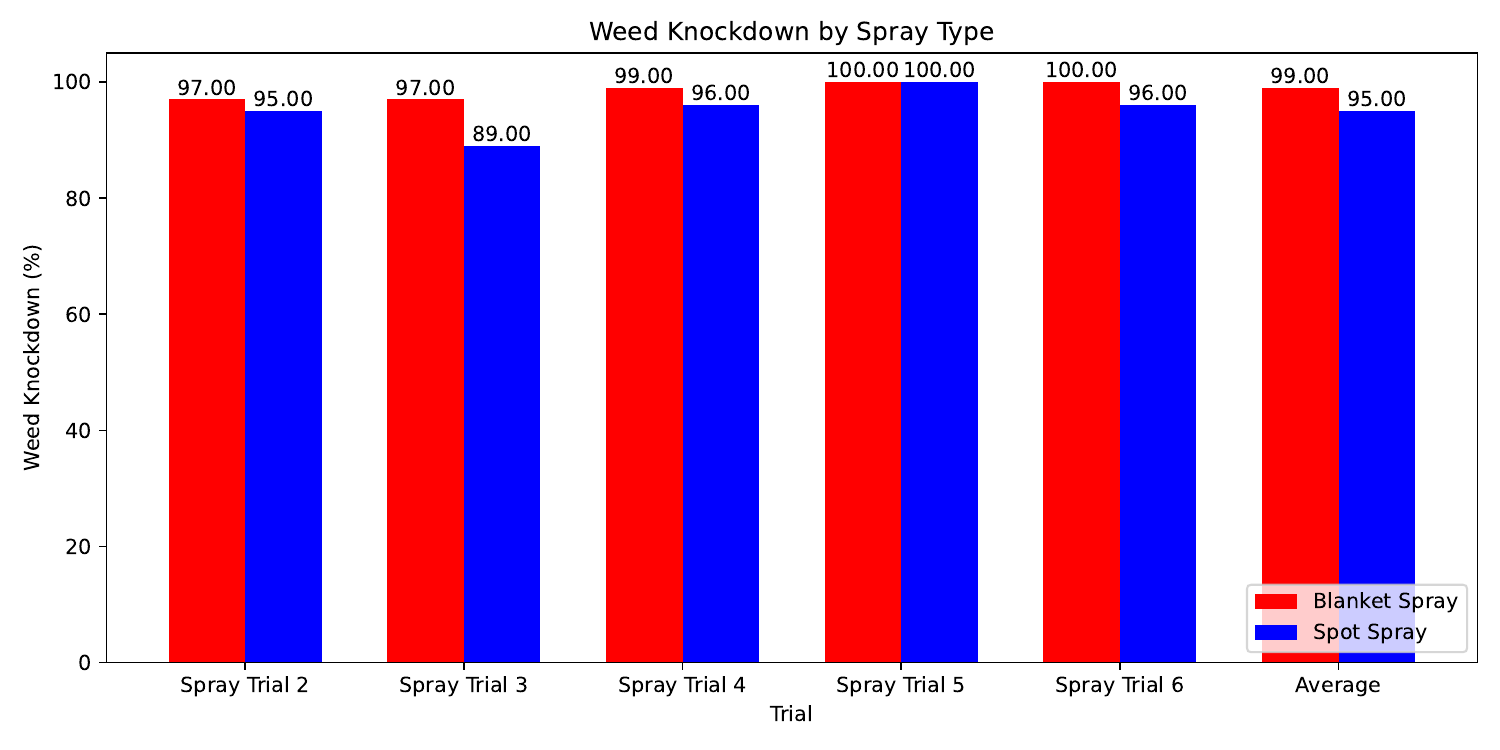}
    \caption{A comparison of the weed knockdown hit rate results (based on Eq. \ref{eq:hitrate}) for blanket spraying and spot spraying and the average hit rate across the six trials.\label{fig:knockdown_chart}}
\end{figure}

Fig.~\ref{fig:knockdown_chart} summarises the weed knockdown hit rate data from Table~\ref{tab:efficacy_results}. This shows that across all trials, robotic spot spraying on sugarcane farms achieved 95\% average weed knockdown compared to 99\% when using the industry standard of blanket spraying. The lowest weed knockdown result for spot spraying was from trial 3, which only achieved an 89\% hit rate compared to 97\% with blanket spraying. Trial 5 achieved the best knockdown with an average of 100\% knockdown across treatments for both spot and blanket spraying. The results for trials 2, 4 and 6 were all comparable to the average result. Another way of comparing spraying efficacy is to use Eq.~\ref{eq:knockdown} that calculates spot spraying hit rate in proportion to that of blanket spraying. The results reported in the second last column of Table \ref{tab:efficacy_results} show that on average across the six trials spot spraying is 97\% as effective as blanket spraying. 

These trials show that spot spraying is not 100\% as effective as the industry practice of blanket spraying. The 4\% drop in weed knockdown hit rate or the 3\% drop in efficacy are largely attributable to instances where the AI detection model failed to classify a target weed. A small percentage of failed classifications are unavoidable when deploying an AI model in a real-world environment with difficult variations. Such unavoidable variations include occlusion of target weeds by the crop plant, or incorrect exposure when the auto-exposure algorithm lags behind a transition of scene brightness. Trial 3 offered the lowest classification results of all trials. A review of the collected images from trial 3 found that a non-trivial number of images were incorrectly under-exposed when shadows from the spraying vehicle were cast on the camera's field of view. This is a potential limitation contributing to the drop in weed knockdown that could be improved in the future. 

Despite the small drop in average performance in these trials, if spot spraying can achieve 95\% weed knockdown hit rate with a large drop in herbicide usage then it is a palatable new tool for sugarcane growers. 

% \begin{figure}[hbt!]
%     \centering
%     \includesvg[inkscapelatex=false, width=0.75\textwidth]{./bar_chart_weed_knockdown.svg}
%     \caption{A comparison of the weed knockdown hit rate results (based on Eq. \ref{eq:hitrate}) for blanket spraying and spot spraying and the average hit rate across the six trials.\label{fig:knockdown_chart}}
% \end{figure}

% 
% \begin{table}[htbp]
% \centering
% \caption{Comprehensive Statistical Analysis of Spray Trial Data}
% \label{tab:spray-trial-analysis}
% \resizebox{\textwidth}{!}{%
% \begin{tabular}{lccccc}
% \hline
% \textbf{Metric} & \multicolumn{2}{c}{\textbf{Spot Spray}} & \multicolumn{2}{c}{\textbf{Blanket Spray}} & \textbf{Test Statistic} \\
%  & Mean & SD & Mean & SD &  \\
% \hline
% \multicolumn{6}{l}{\textit{Overall Summary Statistics}} \\
% Herbicide Used (L) & 25.72 & 0.2524 & - & - & -  \\
% Weed Knockdown (\%) & 95\% & 0.0406 & 99\% & 0.0145 & -   \\
% Herbicide Reduction (\%) & 36\% & - & - & - & -   \\
% \hline
% \multicolumn{6}{l}{\textit{Additional Analyses}} \\
% Correlation: Weed Density vs. Herbicide Used & - & - & - & - & $r = 0.9666^a$   \\
% \hline

% \multicolumn{6}{l}{\small $^a$ Pearson correlation coefficient} \\
% \end{tabular}%
% }
% \end{table}
% 

\subsection{Herbicide usage reduction}
\begin{figure}[hbt!]
    \centering
    \includegraphics[width=0.98\textwidth]{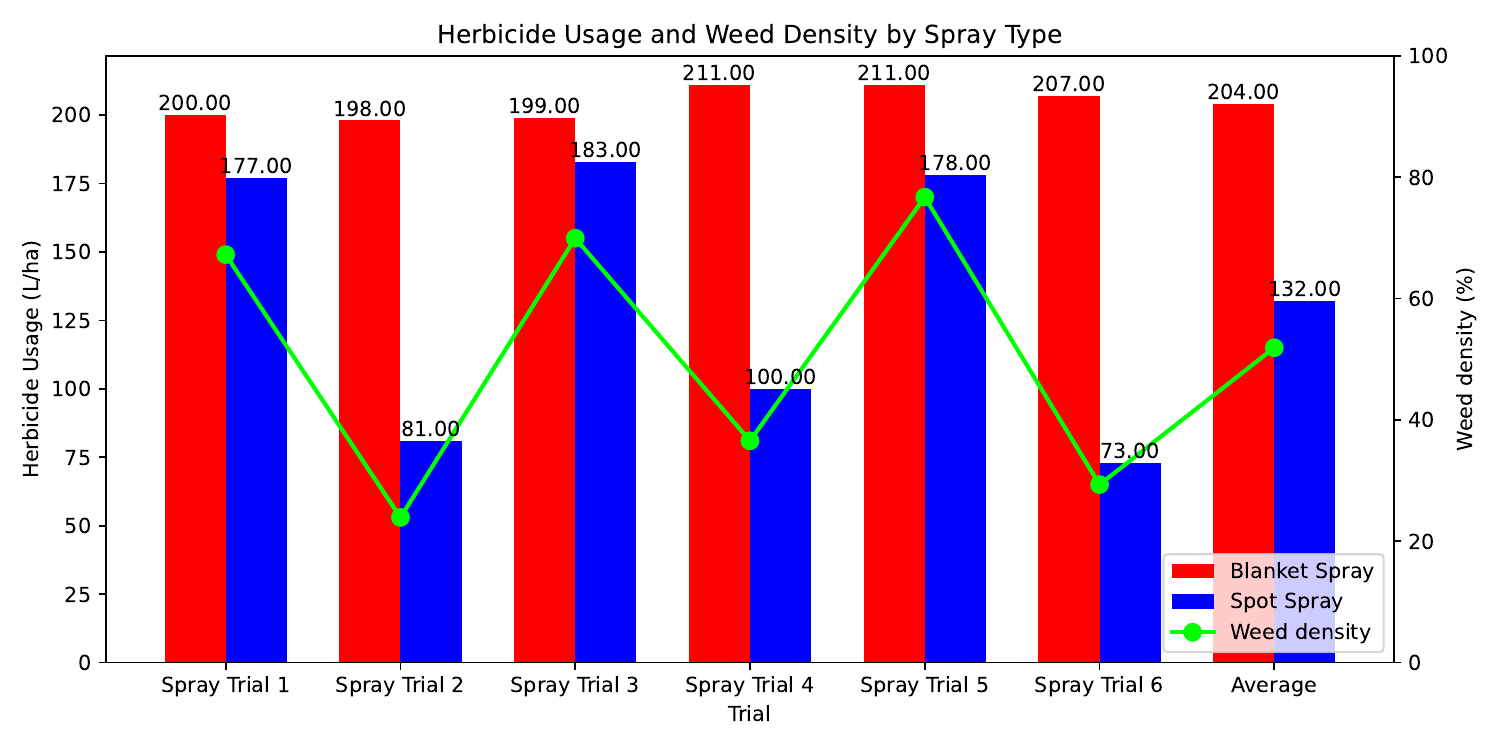}
    \caption{ A comparison of herbicide usage (left axis) for blanket spraying and spot spraying across the trials. Weed density (\%) is measured as the number of images with a weed detected in them divided by the total number of images in each trial area.  \label{fig:herbicide_chart}}
\end{figure}

Herbicide usage was measured for all trials, as shown in Table~\ref{tab:efficacy_results} and is presented in Fig.~\ref{fig:herbicide_chart} averaged across treatments of blanket spraying and spot spraying. Fig.~\ref{fig:herbicide_chart} also superimposes the approximate weed density from each treatment, measured as the number of weed detections divided by the total number of images in each treatment. On average, spot spraying reduced herbicide usage by 35\% across the trials while being 97\% as effective as blanket spraying. And for all trials, the herbicide usage for spot spraying is strongly correlated to the approximate weed density measured in the paddock.

For spray trials 1, 3 and 5, spot spraying used a similar amount of herbicide to blanket spraying, only reducing herbicide usage by 11\%, 8\%, and 16\%, respectively. These trials also recorded strong efficacy results being 98\%, 97\% and 100\% as effective as blanket spraying for trials 1, 3, and 5. The best herbicide reductions were seen for trials 2, 4 and 6, where spot spraying reduced herbicide usage by 59\%, 53\% and 65\%, respectively. Meanwhile, trials 4 and 6 were both 96\% as effective as blanket spraying, but spray trial 2 was only 92\% as effective.

% \begin{figure}[hbt!]
%     \centering
%     \includesvg[inkscapelatex=false, width=0.75\textwidth]{./bar_chart_herbicide_usage_with_weed_density.svg}
%     \caption{A comparison of the herbicide usage results for blanket spraying and spot spraying across the trials (left axis). Along with the approximate weed density of the treated rows (right axis).\label{fig:herbicide_chart}}
% \end{figure}

%Achieving comparable knockdown of weeds with less herbicide has the direct economic benefit of lowering the input costs for herbicide. The Department of Agriculture and Fisheries (DAF) conducted an analysis of the product costs for the herbicides used in this trial work.

%\todo{Ed and Caleb can perhaps add a brief discussion here of their herbicide product cost analysis, including a figure showing the \$/ha cost comparison for blanket spraying and spot spraying across the trials and comment on the results.}

\subsection{Statistical analysis of spray trial data}
This section presents the statistical analysis of the spray trial data, comparing spot and blanket herbicide application methods. The analysis encompasses herbicide usage rates and the correlation between weed density and spot spray application rate. 

\subsubsection{Herbicide usage analysis}
We calculated the mean and standard deviation of herbicide used in L/Ha for both spot and blanket spray methods. This analysis provides insights into the overall herbicide usage and its variability across different plots. As shown in Table \ref{tab:spray-trial-analysis}, the spot-spraying method shows much higher variability (higher Standard Deviation), which is attributed to the highly variable weed density (see Fig. \ref{fig:herbicide_chart}) in the various plots in the 6 trials.  

To determine if there is a significant difference in herbicide usage data between spot and blanket spray methods, we conducted Welch's t-test, which is appropriate for samples with unequal variances. The p-value shows a statistically significant difference between the herbicide usage using two spraying methods.

\subsubsection{Correlation analysis}
Fig. \ref{fig:herbicide_chart} illustrates a clear positive correlation between weed density and herbicide application rate. To quantify this relationship, we calculated the Pearson correlation coefficient, yielding a value of $r = 0.9824$, as shown in Table \ref{tab:spray-trial-analysis}, indicating a very strong correlation between the two variables. This also explains the high standard deviation of herbicide usage in spot spraying trials, as weed density varies randomly across different trial paddocks.  

\begin{table}[htbp]
\centering

\caption{Statistical Analysis of Spray Trial Data}
\label{tab:spray-trial-analysis}
\resizebox{\textwidth}{!}{%
\begin{tabular}{lcccccc}
\hline
\textbf{Metric} & \multicolumn{2}{c}{\textbf{Spot Spray}} & \multicolumn{2}{c}{\textbf{Blanket Spray}} & \textbf{Test Statistic} & \textbf{p-value} \\
 & Mean & SD & Mean & SD & & \\
\hline
%\multicolumn{7}{l}{\textit{Overall Summary Statistics}} \\
Herbicide Usage (L/Ha) & 132 & 50 & 204 & 9 & $t = -4.5754^a$ & 0.0007 \\
%Herbicide Rate (L/ha) & 116.45 & 51.55 & 204.18 & 9.93 & - & - \\
Weed Knockdown (\%) & 94.67 & 4.44 & 98.33 & 1.37 & - & - \\ %$\chi^2 = 3.8889^b$ & 0.0486 
%Herbicide Reduction (\%) & 38.22 & 28.80 & N/A & N/A & - & - \\
%\hline
%\multicolumn{7}{l}{\textit{Trial-by-Trial Analysis}} \\
% Spray Trial 1 (Herbicide Used, L) & 221.5 & 6.36 & 249.5 & 2.12 & - & - \\
% Spray Trial 2 (Herbicide Used, L) & 101.0 & 7.07 & 248.0 & 12.73 & - & - \\
% Spray Trial 2 (Weed Knockdown, \%) & 95.5 & 2.12 & 97.0 & 1.41 & - & - \\
% Spray Trial 3 (Herbicide Used, L) & 206.0 & 7.55 & 220.0 & 0.00 & - & - \\
% Spray Trial 3 (Weed Knockdown, \%) & 89.0 & 7.07 & 97.0 & 1.41 & - & - \\
% Spray Trial 4 (Herbicide Used, L) & 90.0 & 14.14 & 190.0 & 14.14 & - & - \\
% Spray Trial 4 (Weed Knockdown, \%) & 96.0 & 0.00 & 99.5 & 0.71 & - & - \\
% Spray Trial 6 (Herbicide Used, L) & 55.0 & 21.21 & 155.0 & 7.07 & - & - \\
% Spray Trial 6 (Weed Knockdown, \%) & 95.5 & 2.12 & 100.0 & 0.00 & - & - \\
\hline
%\multicolumn{7}{l}{\textit{Additional Analyses}} \\
% \multicolumn{5}{l}{Correlation: Weed Knockdown vs. Herbicide Used} & $r = 0.3824^c$ & - \\
\multicolumn{5}{l}{Correlation: Weed Density vs. Herbicide Used} & $r = 0.9824^b$ & - \\
%\multicolumn{5}{l}{ANOVA: Herbicide Reduction by Crop} & $F = 0.5672^d$ & 0.5894 \\
\hline
\multicolumn{7}{l}{\small $^a$ Paired t-test to compare Herbicide Usage rate between Spot and Blanket methods} \\
%\multicolumn{7}{l}{\small $^b$ Chi-square test for independence between Spray Method and achieving 100\% Weed Knockdown} \\
\multicolumn{7}{l}{\small $^b$ Pearson correlation coefficient} \\
%\multicolumn{7}{l}{\small $^d$ One-way ANOVA F-statistic} \\
\end{tabular}%
}
\end{table}

\subsection{Water quality improvements}

%\todo{Emilie to review this section, which is taken from the WQ reports.}

\begin{figure}[hbt!]
    \centering
    \includegraphics[width=0.98\textwidth]{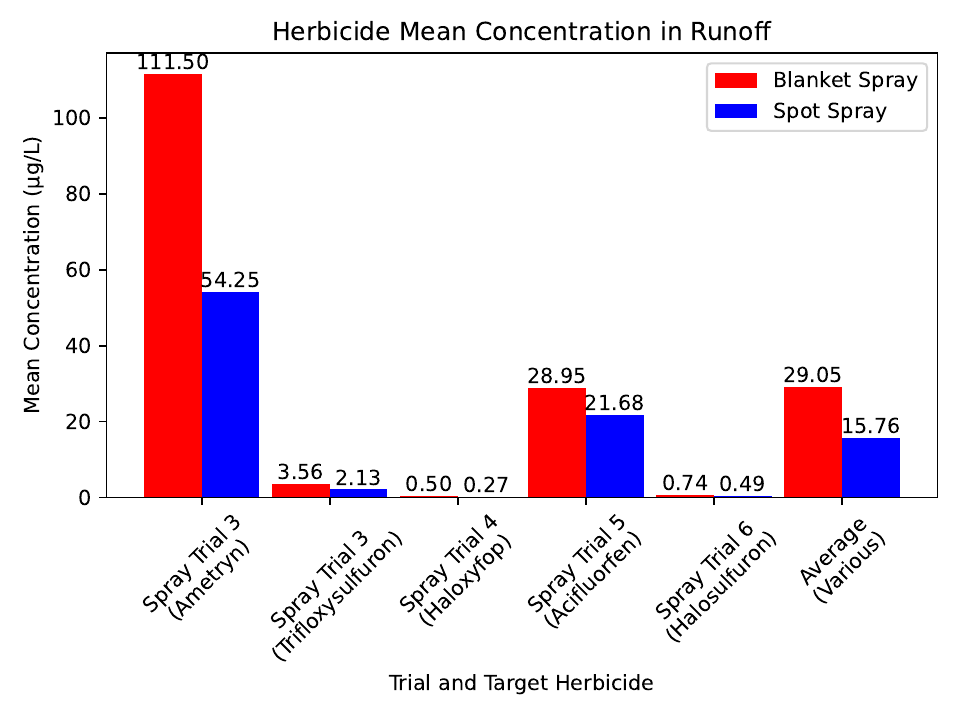}
    \caption{A comparison of the mean concentration of herbicide found in runoff for blanket spraying and spot spraying across the trials.\label{fig:wq_concentration}}
\end{figure}

\begin{figure}[hbt!]
    \centering
    \includegraphics[width=0.98\textwidth]{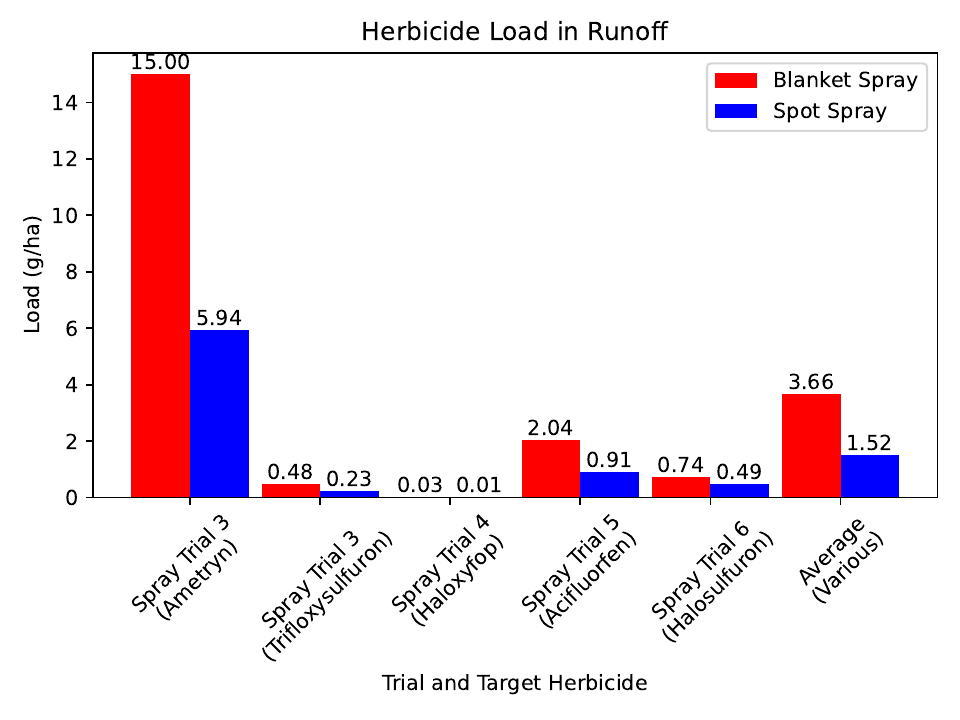}
    \caption{A comparison of the mean herbicide loads found in runoff for blanket spraying and spot spraying across the trials.\label{fig:wq_loads}}
\end{figure}

Water quality sampling and analysis was performed following the aforementioned methodology for trials 3, 4, 5, and 6. Table~\ref{tab:wq_results} documents the results and Figs.~\ref{fig:wq_concentration} and~\ref{fig:wq_loads} compare the mean concentrations and loads of herbicide found in runoff between spot spraying and blanket spraying. 

Figs.~\ref{fig:wq_concentration} and~\ref{fig:wq_loads} show that herbicide concentrations and loads are proportional to the amount of herbicide sprayed across all trials. They also show that on average across all trials, spot spraying reduced the mean concentrations and loads of herbicides in runoff by 39\% and 54\% of the blanket spray, respectively.
 By limiting the amount of herbicide sprayed to areas with weed coverage and not spraying bare ground, spot spraying technology can reduce the environmental footprint of herbicide control.

% \begin{figure}[hbt!]
%     \centering
%     \includesvg[inkscapelatex=false, width=0.75\textwidth]{./bar_chart_wq_concentration.svg}
%     \caption{A comparison of the mean concentration of herbicide found in runoff for blanket spraying and spot spraying across the trials.\label{fig:wq_concentration}}
% \end{figure}

% \begin{figure}[hbt!]
%     \centering
%     \includesvg[inkscapelatex=false, width=0.75\textwidth]{./bar_chart_wq_loads.svg}
%     \caption{A comparison of the mean herbicide loads found in runoff for blanket spraying and spot spraying across the trials.\label{fig:wq_loads}}
% \end{figure}

Spray trial 3 included separate analyses for the two active ingredients of the Krismat herbicide product. Much higher loads and concentrations of ametryn were found in runoff compared to trifloxysulfuron because of its higher concentration in the product that was applied. For both active ingredients in spray trial 3, concentrations and loads were directly proportional to the amount of herbicide applied. Ametryn is reportedly very toxic at low concentrations to aquatic plants and algae (i.e. EC50 algae acute 72 hour – 0.0036 mg/L).  

For spray trials 4 and 5, the concentrations and loads were generally proportional to the amount of herbicide applied. Despite very low concentrations found in the runoff, haloxyfop and acifluorfen only have low to moderate impact on aquatic organisms (EC50 aquatic plants acute 7 day – 0.0002 mg/L, EC50 algae acute 72 hour – 0.0053 mg/L).

Before spray trial 6, ametryn was blanket sprayed through the whole trial block. Halosulfuron load and concentration runoff data were, therefore, expressed in proportion to ametryn data (not shown). Similar to previous trials, halosulfuron concentrations and loads were proportional to the amount of herbicide sprayed, but generally higher. 

\color{blue} 
Paired t-tests on trial averages indicate a trend of reduced knockdown (Fig. \ref{fig:knockdown_chart}), concentration (Fig. \ref{fig:wq_concentration}), and loads (Fig. \ref{fig:wq_loads}) with spot spraying, aligning with expectations. However, the small sample sizes and small number of within-trial replicates limit statistical power.
\color{black} 

\section{Discussions and Future Works} \label{Sec:Dis}

In this study, we report a comprehensive set of field experiments testing robotic weed spot spraying when targeting different weeds and using different chemical herbicides. The results are very promising in terms of maintaining a high weed knockdown efficacy whilst significantly reducing the amount of herbicide applied and consequently, the amount of herbicide detected in water run-off induced by irrigation. Therefore, robotic spot spraying has substantial potential to reduce herbicide use.

The technology presented in this project was specifically developed for managing sugarcane weeds. However, crop rotation in sugarcane paddocks is a fundamental practice in sustainable agriculture, contributing to long-term soil productivity, environmental health, and economic viability for farmers. We demonstrated that our technology is effective even when different rotation crops are cultivated. This adaptability is a strength of our spot-spraying technology that can seamlessly adjust to other weeds and crops using the same detection and spraying hardware, as well as the same detection framework. Our DL detection method remains consistent across the five different paddocks (with varying growth and natural environments), three different weed categories (with diverse morphologies, colours, and textures), and the two different crops (with distinct morphologies, colours, and textures). Additionally, our trial 6 was conducted in a plant cane rather than a ratoon cane, adding extra variability to our experiments. This approach is aligned with the well-known site-specific approach, one needs to retrain a previous model or develop a new model using data from a new paddock, which could have a previously addressed or a new weed/crop.

The scope of this work was limited to sugarcane (and a rotation crop in sugarcane) paddocks with common weeds in the Burdekin region of Queensland, Australia. Therefore, the datasets and detection models collected and developed herein may not have utility in other sugarcane farming regions, and a site-specific approach should be adopted for new weed/crop scenarios.  Nevertheless, we expect that the general conclusions regarding the effectiveness of the technology would apply to other regions.

Another limitation of our work is the lack of yield data for our trials. This was primarily because yield analysis was beyond the scope of the current study, as we did not have the resources to obtain and analyse yield data for the farm blocks where we conducted our spraying trials. However, it is known that registered herbicides can negatively impact crop yield. Sugar Research Australia routinely screens a range of herbicides for their impact on cane varieties, with results updated yearly and available in the Sugar Research Australia variety guides \cite{sugar_research_australia_2024}. When weeds are controlled and not competing with the crop, reduced herbicide usage may improve yield. Since we have demonstrated that spot-spraying with AutoWeed can reduce herbicide use while maintaining weed control, we expect that AutoWeed may positively impact overall yield. This needs to be fully verified in future studies.

Another limitation of this work is that it heavily relies on human annotation of collected datasets in order to train weed detection models. This study does not take advantage of recent innovations in semi-supervised and unsupervised learning approaches. Such approaches may drastically increase the speed of annotation and turnaround time from dataset collection to spraying. In order to build larger datasets with regional invariance, unsupervised and semi-supervised methods must be relied upon.

This work is also limited to targeting nutgrass in ratoon sugarcane, nutgrass in plant sugarcane and grass and broadleaf weeds in mung beans. There are many other priority weeds of interest to the Burdekin region and other sugarcane farming regions in Australia and internationally. Some challenging weed species exist which present as visually similar to sugarcane, such as Guinea grass and wild sorghum.

Another limitation of our study is the slight unintentional variations in our experiments, which occurred due to the nature of trial work involving several uncontrollable factors. Ideally, we would have conducted all trials on sugarcane. However, only four out of six trials were performed on sugarcane, with the remaining two conducted on a rotation crop on a sugarcane farm to utilise all available opportunities. Additionally, we had to use two different spraying systems due to the unavailability of the initial farming machine. Nonetheless, this did not impact our results as the detection and spraying control electronics remained consistent.

Future work in this area includes: investigating and applying shadow removal techniques for improved weed classification in real-world situations, removing or lessening the burden of human annotation with unsupervised or semi-supervised learning approaches, targeting hard-to-identify perennial grass weed species which present as visually similar to sugarcane crops, and building larger datasets and training AI models that are robust to regional variance in weeds and sugarcane crops outside the Burdekin region, within which this work was limited.

\section{Conclusions}\label{Sec:Conc}
This study highlights how advanced robotics combined with computer vision and deep learning can revolutionise weed control in agriculture. The field evaluation of AutoWeed shows that precision application not only reduces herbicide usage on sugarcane farms but also promotes a more sustainable approach by minimising chemical use.

By targeting only the necessary areas, AutoWeed shifts the focus from blanket application to a smarter method, cutting input costs and reducing the chemical burden on ecosystems. The resultant improvements in water quality further underscore the environmental benefits of this technology.

In summary, AutoWeed represents a significant step forward in precision agriculture, setting the stage for future innovations that marry technology with sustainability to safeguard both farm productivity and the environment.
\color{black}
%In conclusion, this work demonstrates the efficacy of a custom-designed robotic spot spraying tool, AutoWeed, in reducing herbicide usage on sugarcane farms while maintaining effective weed control. Our field trials, spanning 25 hectares of sugarcane farms, show that AutoWeed spot spraying is 97\% as effective at controlling weeds compared to the industry-standard broadcast spraying. Importantly, it also offers a substantial reduction in herbicide usage, with an average reduction of 35\%, directly correlated with weed density. For areas with lower weed pressure, the reduction in herbicide usage can be as high as 65\%. Our analysis of irrigation-induced runoff also revealed that spot spraying reduced both the mean concentration and mean load of herbicides in runoff by 39\% and 54\%, respectively, compared to broadcast spraying. These findings underscore the potential of precise robotic spot spraying to benefit both farmers and the environment, by cutting input costs and providing sustained water quality improvements.

\section{CRediT authorship contribution statement}

Mostafa Rahimi Azghadi: Conceptualisation, Methodology, Data Collection, Data Curation, Data Analysis, Funding Acquisition, Writing Original Draft, Supervision, Project administration. Alex Olsen: Conceptualisation, Methodology, Software, Data Collection, Data Curation, Data Analysis, Visualization, Funding Acquisition, Writing Original Draft, Supervision, Project administration. Alzayat Saleh: Data Analysis, Visualization, Writing – review \& editing.  Jake Wood: Conceptualisation, Methodology, Software, Data Collection, Data Curation, Data Analysis, Visualization. Brendan Calvert: Methodology, Data Collection, Data Curation. Terry Granshaw: Conceptualisation, Methodology, Data Collection, Data Analysis. Emilie Fillols: Conceptualisation, Methodology, Data Collection, Data Analysis, Reviewing/editing the draft. Bronson Philippa: Conceptualisation, Methodology, Data Analysis, Reviewing/editing the draft.

\section{Acknowledgement}
This research is funded by the partnership between the Australian Government's Reef Trust and the Great Barrier Reef Foundation. We thank the participating farmers for allowing us to carry out our experiments on their properties. 

%\appendix

%\section{Sample Appendix Section}
%Lorem ipsum dolor sit amet, consectetur adipiscing elit, sed do eiusmod tempor section \ref{sec:sample1} incididunt ut labore et dolore magna aliqua. Ut enim ad minim veniam, quis nostrud exercitation ullamco laboris nisi ut aliquip ex ea commodo consequat. Duis aute irure dolor in reprehenderit in voluptate velit esse cillum dolore eu fugiat nulla pariatur. Excepteur sint occaecat cupidatat non proident, sunt in culpa qui officia deserunt mollit anim id est laborum.

 \bibliographystyle{elsarticle-num} 
 \bibliography{cas-refs}





% \end{thebibliography}
\end{document}